\documentclass[11pt, a4paper, logo, twocolumn, copyright]{deepmind}

\pdftrailerid{redacted}

\makeatletter
\renewcommand\bibentry[1]{\nocite{#1}{\frenchspacing\@nameuse{BR@r@#1\@extra@b@citeb}}}
\makeatother

\usepackage{kantlipsum, lipsum}
\usepackage{dsfont}
\usepackage{dm-colors}
\usepackage[ruled,vlined,linesnumbered]{algorithm2e}

\usepackage[authoryear, sort&compress, round]{natbib} 
\usepackage{subcaption}

\usepackage{amsthm}
\newtheorem{definition}{Definition}

\usepackage{listings}
\graphicspath{{figures/}}

\title{Faithful Reasoning Using Large Language Models}

\correspondingauthor{tonicreswell@deepmind.com}


\renewcommand{\today}{August 2022}

\author[1]{Antonia Creswell}
\author[1]{Murray Shanahan}

\affil[1]{DeepMind}

\begin{abstract}

Although contemporary large language models (LMs) demonstrate impressive question-answering capabilities, their answers are typically the product of a single call to the model. This entails an unwelcome degree of opacity and compromises performance, especially on problems that are inherently multi-step. To address these limitations, we show how LMs can be made to perform {\em faithful} multi-step reasoning via a process whose causal structure mirrors the underlying logical structure of the problem. Our approach works by chaining together reasoning steps, where each step results from calls to two fine-tuned LMs, one for selection and one for inference, to produce a valid reasoning trace. Our method carries out a beam search through the space of reasoning traces to improve reasoning quality. We demonstrate the effectiveness of our model on multi-step logical deduction and scientific question-answering, showing that it outperforms baselines on final answer accuracy, and generates humanly interpretable reasoning traces whose validity can be checked by the user.
\end{abstract}

\begin{document}

\maketitle

\section{Introduction}

Among the many tasks that contemporary large language models (LMs) can perform \citep{zeng2022socratic, nakano2021webgpt, alayrac2022flamingo}, question-answering is potentially one of the most useful \citep{rae2021scaling}. However, the proficiency of these models typically goes hand-in-hand with an unacceptable level of opacity. The assumptions behind an answer and the intermediate steps of reasoning that justify it -- insofar as these exist -- are hidden from the user. This prevents the user from verifying an answer, makes it difficult to debug a model when it gets an answer wrong, and undermines overall trust in the model’s responses.

\begin{figure*}[h!]
    \centering
    \includegraphics[width=0.8\textwidth]{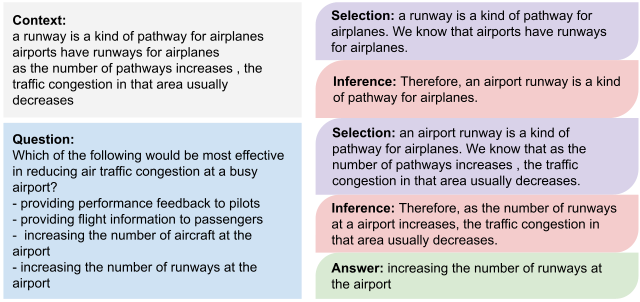}
    \caption{\textbf{Example input and output from our Faithful Reasoning model.}}
    \label{fig:top_example}
\end{figure*}

By contrast, a system that reasons {\em faithfully} is one whose underlying computations mirror standard definitions of logical validity. Such a system can supply the user with an interpretable reasoning trace, which allows them to understand how the model reached its final answer. Exposing a model’s assumptions and reasoning steps (Figure \ref{fig:top_example}) in this way enables the user to spot mistakes the model may have made, and empowers them to decide for themselves whether the model's conclusions are justified.

This provision is especially important given that LMs are trained on human data collected from the internet, which makes them vulnerable to picking up and perpetuating bias \citep{betz2020critical, bender2021stochastic, weidinger2021ethical}. Presented with a context of relevant knowledge and a question, an LM may base its answer on information encoded in its weights rather than prioritising the information present in the context \citep{dasgupta2022language}. Without an interpretable reasoning trace, we cannot know how a model has reached its answer. Did the model rely on its priors, which may be biased, or did it obtain an answer by reasoning correctly with relevant knowledge?

\begin{figure*}[h]
    \centering
    \includegraphics[width=0.8\textwidth]{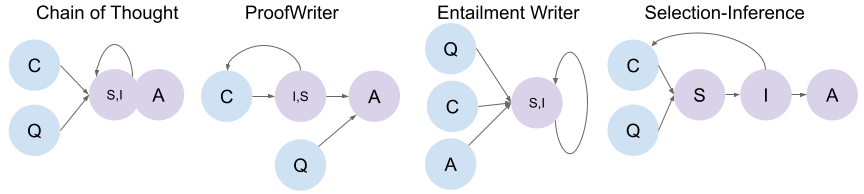}
    \caption{\textbf{Comparing dependencies between inputs and outputs for SI and related models.} Inputs - blue circles, LM outputs - purple circles. Order of the letters indicates the order in which the values are predicted. Arrows indicate the dependencies between inputs -- the context, C, and question, Q -- intermediate outputs -- the selection, S, and inference, I, and the final answer, A. SI is the only model where the answer does not have a direct dependency on the question. Note, EntailmentWriter takes the hypothesis and context as input, where the hypothesis depends on the question and answer.}
    \label{fig:compare_causal}
\end{figure*}

In this paper we develop a forward-chaining model that reasons faithfully in the sense defined above (and more formally in Section \ref{sec:defn}). The backbone of our system, denoted SI, comprises two fine-tuned LMs, one for {\em selection} and one for {\em inference}. The interleaved operation of these two components has a causal structure (Figure \ref{fig:compare_causal}) that mirrors the definition of logical validity. This guarantees that the model's answers follow logically from the given context under certain assumptions.

Two further fine-tuned language models complete our architecture. First, the {\em halter} is used to terminate the reasoning process and return an answer in the required format. If the trace does not terminate within a specified number of steps then the answer is considered to be {\em `Unknown'}, allowing us to filter model answers and increase answer precision. Second, a learned {\em value function}, which assesses the quality of the current reasoning step is deployed to guide a beam search over reasoning traces to enhance their quality and further boost overall performance.

We evaluate our model on two datasets, ProofWriter \citep{tafjord2020proofwriter} and a question-answering version of EntailmentBank \citep{Dalvi2021ExplainingAW}. We shown that our model outperforms baseline models on final answer accuracy and that our proposed halter and search methods also lead to compounding boosts in performance (Tables \ref{tab:main_eb_results} and \ref{tab:main_pw_results}). We show that in most cases SI produces higher quality reasoning traces than baseline models  (Figures \ref{fig:eb_proof_eval} and \ref{fig:pw_proof_eval}). It is less likely to ``hallucinate'' facts (Table \ref{tab:eb_made_up}), is better able to utilise the context (Table \ref{tab:eb_random_context}) and is more likely to use its trace to answer questions (Table \ref{tab:pw_incomplete_context}). Finally, our model can accurately predict when it knows the answer (Figure \ref{fig:all_vs_knowns}).

\begin{figure}[h]
    \centering
    \includegraphics[width=0.9\columnwidth]{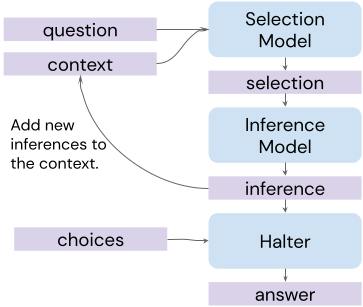}
    \caption{\textbf{Faithful Reasoning architecture.} See Section \ref{sec:selection_inference} for details of each component.}
    \label{fig:compare}
\end{figure}

\section{Defining a Valid Reasoning Trace}\label{sec:defn}

In this Section, we formally define the concept of \textit{valid} forward reasoning in the context of our framework, adhering closely to textbook definitions from formal logic (e.g. \cite{hamilton1988logic}).

\begin{definition} \label{defn:step}
A {\em reasoning step} is a pair $\langle s, i \rangle$, where $s$ (the selection) is a set of statements and $i$ (the inference) is a statement.
\end{definition}

\begin{definition}\label{defn:trace}
A {\em reasoning trace} is a pair $ \langle \mathcal{C}, \mathcal{T} \rangle$ where $\mathcal{C}$ (the context) is a set of statements and $\mathcal{T}$ is a sequence of reasoning steps.
\end{definition}

\begin{definition}\label{defn:connected}
A reasoning trace $ \langle \mathcal{C}, \mathcal{T} \rangle $, where $\mathcal{T} = \langle s_0, i_0 \rangle, \langle s_1, i_1 \rangle, … \langle s_n, i_n \rangle$, is {\em connected} iff for every reasoning step $\langle s_k, i_k \rangle$, for every statement $q$ in the set $s_k$ either $q \in \mathcal{C}$ or $q = i_j$ for some $j < k$.
\end{definition}

\begin{definition}\label{defn:valid}
A reasoning trace $\langle \mathcal{C}, \mathcal{T} \rangle$, where $\mathcal{T} = r_0, r_1, …, r_n$,  is {\em valid} if it is connected and each reasoning step $r_k = \langle s, i \rangle$ is correct (in the sense that $i$ logically follows from $s$).
\end{definition}

In the next section, we introduce the components of our architecture and show how it satisfies the requirements of faithful reasoning, under certain assumptions.

\section{Components of a Faithful Reasoning Model}\label{sec:selection_inference}

We begin by introducing Selection-Inference (SI), the step-wise forward reasoning backbone whose causal structure (see Figure \ref{fig:compare_causal}) satisfies the requirements for producing valid reasoning traces. We then describe a component for \textit{halting}, which looks at the output of a Selection-Inference step and determines if there is sufficient information to answer the question. When there is sufficient information, the model predicts the answer in such a way that it cannot rely on knowledge embedded in its weights, but must depend on the reasoning trace. Finally, we introduce a \textit{value function}, which is used to perform a step-level beam search on the reasoning traces to find the best candidate for answering the question. A schematic of our model is shown in Figure \ref{fig:compare}. We now describe each of these components in more detail. Note that (in contrast to \cite{Dalvi2021ExplainingAW}) each component in our model is trained in isolation, and at no point do we optimise our pipeline for final answer accuracy.

\subsection{Selection-Inference: Valid Forward Reasoning}

Given a question and a context consisting of a number of statements sufficient to answer the question, we would like our model to produce a sequence of deductive reasoning steps that answers the question (Figure \ref{fig:top_example}). To achieve this, the SI backbone splits each reasoning step in two (Defn. \ref{defn:step}). First, given the question, the Selection model chooses a set of statements from the context (the {\em selection}). Second, the Inference model predicts an entailment by computing a statement that follows from the selection (the {\em inference}). The inference is then added to the context, and that concludes a single step of reasoning. Multiple iterations of SI are carried out to produce a reasoning trace (Defn. \ref{defn:trace}). The final inference is used to answer the question.

\subsubsection{Selection}\label{sec:selection}

\begin{figure*}[h]
    \centering
    \includegraphics[width=0.65\textwidth]{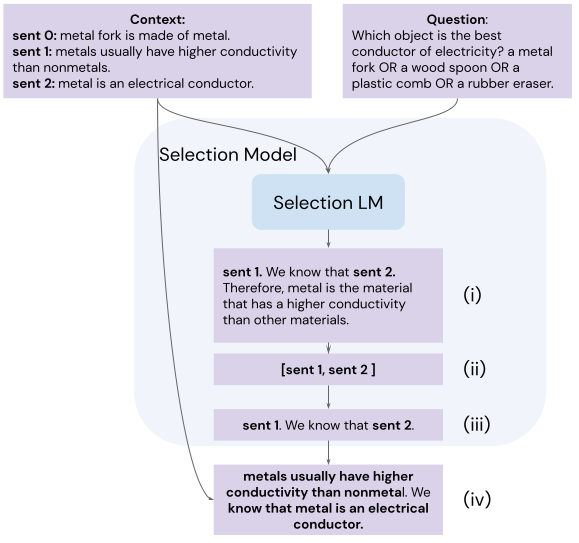}
    \caption{\textbf{The Selection model.} The role of the Selection model is to take the context and question and select a number of statements from the context to feed to the inference model. It is crucial that the Selection model is not able to `hallucinate` facts. To achieve this we fine-tune a LM to predict sentence labels, as show in (i). We then extract the only the sentence labels (ii) and compose a sentence (iii). The statements from the context are then substituted back in (iv) resulting in a sentence composed of statements from the context.}
    \label{fig:selection}
\end{figure*}

To ensure that the reasoning trace is \textit{connected} (Defn. \ref{defn:connected}), the Selection model is obliged to select elements only from the context, and is unable to `hallucinate' facts. Similar to \cite{tafjord2020proofwriter} and \cite{Dalvi2021ExplainingAW}, we achieve this by training an LM to \textit{refer} to statements in the context by their \textit{sentence labels}, for example, `sent 3'. These are used to compose sentences of the form ``X. We know that Y and ... and Z.'', where X, Y, and Z are sentence labels (Figure \ref{fig:selection}). These sentences are passed directly to the inference model.

\subsubsection{Inference}
To encourage it to produce \textit{correct} reasoning steps, the Inference model is trained to predict an entailment given {\em only} the selection. By not allowing the Inference model access to the question, we prevent it from ``cheating'' (directly predicting the answer from the question). While we cannot guarantee that every reasoning step is correct, in the sense that the inference logically follows from the selection (Defn. \ref{defn:valid}), our implementation makes this more likely. Under the assumption that the Inference model produces logically correct inferences, our model is guaranteed to produce valid reasoning traces.

\subsection{Halting: When to Stop Reasoning?}\label{sec:halt}

SI allows us to produce multi-step reasoning traces, but it does not tell us when to stop the reasoning process. Furthermore, while we may want to use the final inference as the answer, it may not be in a desirable format. For example, if we are asked whether a statement `$P(X)$' is true or false, our final inference my be `$P(X)$' or `$not P(X)$' where $P$ is a predicate and $X$ a constant. Alternatively, we may want to answer multiple-choice questions, which require one answer to be output from a given set of possibilities.

\begin{figure*}[h]
    \centering
    \includegraphics[width=\textwidth]{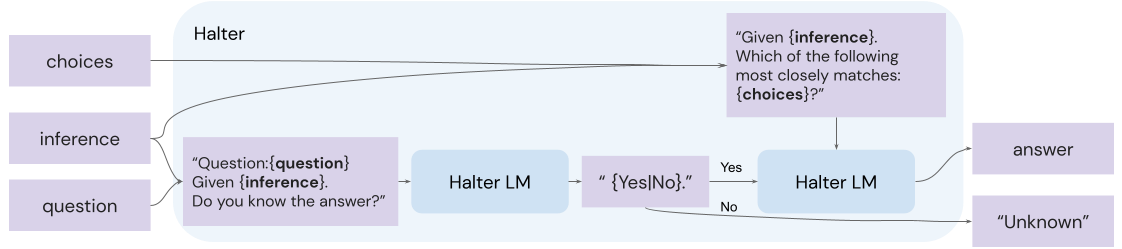}
    \caption{\textbf{The two-stage Halter.} First the model determines if a question is answerable given the current inference. If it is, the model combines minimal additional information (that could not be used on its own to answer the question) and predicts the answer.}
    \label{fig:halter}
\end{figure*}

In light of this, we deploy a two-stage Halter (Figure \ref{fig:halter}), which uses an LM fine-tuned to predict whether the question can be answered given the current inference and the question. If the question cannot be answered, \textit{`Unknown'} is returned. Otherwise, the Halter computes an answer, using the same LM, given the final inference and minimal additional information. It is important that the model is obliged to \textit{use} the final inference, rather than depend on knowledge embedded in its weights. For example, if we are answering a multiple-choice question, we may provide the choices alongside the final inference, and use the model to output the choice that most closely matches that inference.

To determine if the system is ready to answer the question, we provide the Halter with a sentence of the following form: \textit{`Question:\{question\} Given \{inference\}. Do you know the answer?'}. The output of the Halter LM is then either \textit{`Yes'} or \textit{`No'}. If the output is \textit{`Yes'}, the Halter LM is then prompted {\em again} to answer the question with a prompt of the following form: \textit{`Given \{inference\}. Which of the following most closely matches: \{choices\}? Answer:'}. The output is one of the choices.

The Halter is applied after each step of SI to the resulting inference. If the output of the Halter is \textit{`Unknown'} then we proceed to another iteration of SI. If the output of the Halter is an answer, then the process is terminated and the answer is returned. If, after a pre-specified number of SI iterations, the system has not halted, it returns the answer \textit{`Unknown'} (Alg. \ref{alg:halt}). An additional benefit of this is that it allows the model to say that it cannot answer the question, rather than making up an answer. We see a notable increase in performance when we remove questions that the model ``thinks'' it cannot answer (Figure \ref{fig:all_vs_knowns}). This has significant implications for trust, safety and the deployment of systems in the real world, where precision (rather than recall) is a priority.

\subsection{Search: Finding the \textit{Best} Trace}\label{sec:search}

The selection module is non-deterministic, in the sense that it samples from multiple candidate statements, and this induces a tree of potential reasoning traces. We use beam search to explore this tree in order to find high quality traces. To enable this, we introduce a \textit{value function} which computes the value of adding a reasoning step to the current trace. The value function is a language model, $\text{LM}_{\text{value}}$, fine-tuned on examples of {\em partial} reasoning traces that culminate in a ``correct'' or ``incorrect'' next step. A step is considered ``correct'' if it is both logically valid and is on the ground truth (shortest) reasoning path. A step is otherwise considered ``incorrect''.

Assuming that the sum of probabilities for ``correct'' and ``incorrect'' is close to one, we can use $\log p_{\text{value}}(\text{``correct''}|\text{reasoning trace})$ to score reasoning traces as they are being constructed, where $log p_{\text{value}}$ denotes the distribution over tokens learned by the language model, $\text{LM}_{\text{value}}$.

We use the value function to guide a beam search. Starting from a single empty trace we use SI to produce $P$ candidate steps. We evaluate each of these steps using the value function and keep the top $B <= P$. We use SI again to generate $P$ candidate next steps for each of the $B$ traces, resulting in $B\times P$ traces. These are evaluated using the value function and the best $B$ traces are kept. We continue this process until all the traces have halted.

\section{Experimental Setup and Evaluation of Components}

In this section, we detail how each component in our faithful reasoning model is trained, and evaluate each component is isolation, where possible. We use two challenging reasoning datasets, ProofWriter \citep{tafjord2020proofwriter} and a modified -- more challenging -- question-answering version of EntailmentBank \citep{Dalvi2021ExplainingAW}. We use a 7B parameter Chinchilla language model in each of our components \citep{hoffmann2022training}.

\subsection{Datasets}

We fine-tune language models on examples of ground truth reasoning traces. Two datasets that provide reasoning traces are EntailmentBank \citep{Dalvi2021ExplainingAW} and ProofWriter (PW) \citep{tafjord2020proofwriter} (See Section \ref{sec:proofwriter_sup} for details). ProofWriter is a dataset of logical reasoning problems that ask a question whose answer is True or False given a context and provide step-by-step reasoning traces. Problems require 1, 2, 3 or 5 steps of reasoning. EntailmentBank is derived from the ARC \citep{clark2018think} dataset of \textit{grade school} science questions. \cite{Dalvi2021ExplainingAW} provide a dataset of $\langle$context, hypothesis, entailment tree$\rangle$ triples. \cite{Dalvi2021ExplainingAW} propose three tasks, \textit{Task 1} where the context consists of facts from WorldTreeV2 \citep{xie2020worldtree} needed to answer the question, and \textit{Task 2} that additionally includes distractors. EntailmentBank is not a QA dataset, rather the task requires predicting the entailment tree given the hypothesis and context. We reformulate the EntailmentBank dataset (taking additional information from the original ARC tasks) into an EntailmentBankQA (EB) dataset by creating a dataset of context, question, choices, answer and a proof derived from the entailment tree. Our task is to predict the answer and proof given the question, context and choices. This task is more similar to the ARC task, however, here we provide the context and predict a reasoning trace that leads to the answer.

\subsection{Selection-Inference}

The selection model is trained on individual steps of reasoning; given the context and any previous inferences the model is trained to predict the sentence labels which refer to statements in the context. The inference model is trained to predict an entailment given a number of statements from the context. Each reasoning step in each training example in the original dataset produces one training data point for selection and one training point for inference. Examples of $\langle$input, target$\rangle$ pairs used to train the LMs are shown in Figures \ref{fig:pw_si_train} and \ref{fig:eb_si_train}. By training the model to select statements by labels we prevent the model from being able to make up facts that are not present in the context (Tables \ref{tab:pw_made_up} and Table \ref{tab:eb_made_up}). Tables \ref{tab:pw_inference} and \ref{tab:eb_inference} show the inference accuracy on the test set.

\subsection{Halter}

We use the ground truth reasoning traces from each dataset to produce training examples for the Halter LM. The halter has two functions, (1) learn when there is sufficient information to answer the question given the current inference (and the question) and (2) answer the question given the current inference and the choices. An example of how data is generated is shown in Figure \ref{fig:halter_train}. Each step of reasoning in each problem can be converted into a data point for training. The input has the form \textit{`Question: \{question\}. Given \{inference\}. Do you know the answer?'}. For intermediate reasoning steps the target is \textit{` No.'}. For final reasoning steps the target is \textit{` Yes.'}. From these examples, the model can learn whether an inference contains sufficient information to answer the question. We obtain an additional data point for each problem, which is used to train the model to answer a question. The inputs are of the form \textit{`Given \{inference\}. Which of these most closely matches \{choices\}?'}. The target is the ground truth answer given in the dataset.

We train two halters, one on the PW dataset and the another on the EB dataset. For the PW dataset we use a simplified single step prediction because the question does not contain sufficient information to solve the problem\footnote{Note that for every "Show P(X)" there is a "Show not P(X)" in the PW dataset, therefore if the model tried to answer using only the question the model would achieve only 50\% accuracy.}. Specifically, for PW we construct a training dataset where the input has the form \textit{`Given \{inference\}. \{question\}'}. For each intermediate inference the target is \textit{` Unknown'} while the final inference has the target \textit{` True'} or \textit{` False'}.

To evaluate each halter independently of the \textbf{Proof Only} baseline or SI model, it is applied it to the ground truth proofs from the test split. Tables \ref{tab:main_pw_results} and \ref{tab:main_eb_results} show results for PW and EB respectively. We see that the PW halter performs almost perfectly while the EB halter achieves $88.8\%$ accuracy.

\begin{figure*}[h]
\centering
    \begin{subfigure}[b]{0.4\textwidth}
        \centering
        \includegraphics[width=\textwidth]{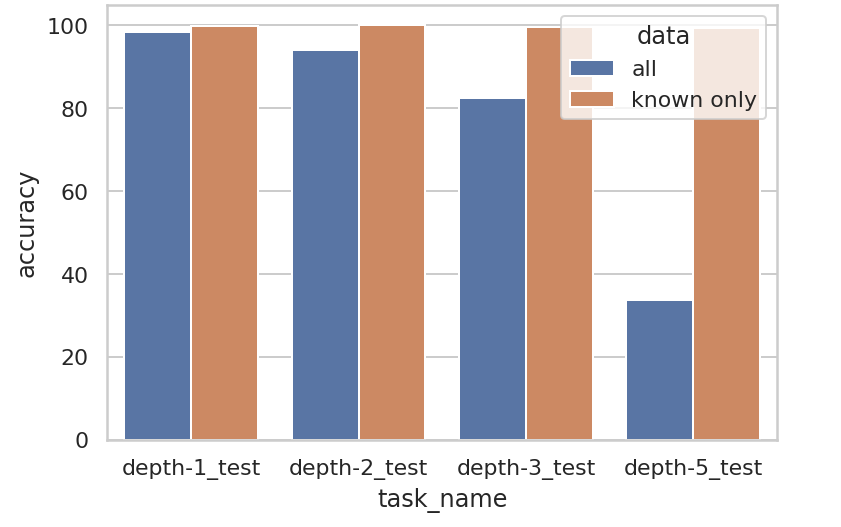}
        \caption{\textbf{ProofWriter}}
        \label{fig:pw_all_vs_knowns}
    \end{subfigure}\hspace{10mm}%
    \begin{subfigure}[b]{0.4\textwidth}
        \centering
        \includegraphics[width=\textwidth]{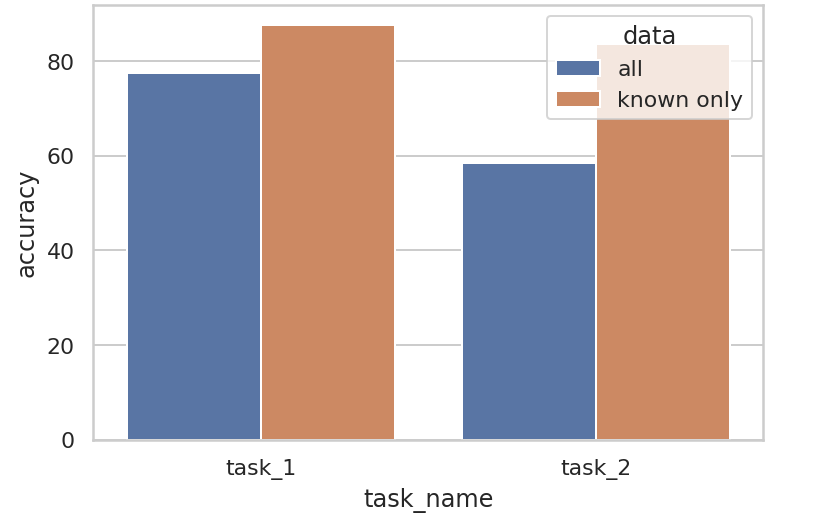}
        \caption{\textbf{EntailmentBankQA}}
        \label{fig:eb_all_vs_knowns}
    \end{subfigure}
    \caption{\textbf{Our model accurately predicts when it `knows' the answer}. The `known only' accuracy is computed after filtering out the answers that are `Unknown'. The `all' accuracy is computed on all problems. This property is beneficial for applications that require high precision.}\label{fig:all_vs_knowns}
\end{figure*}

The Halter endows our model with the desirable property of predicting when it \textit{does not know the answer}. Figure \ref{fig:all_vs_knowns} shows that our halter model can reliably predict when the answer is known. When we filter out the problems where the model does not know the answer, we obtain nearly perfect accuracy on the PW dataset for all depths and 87.5\% \& 83.7\% accuracy on Task 1 and 2 of EB dataset respectively. This has significant implications for the deployment of such models in scenarios where precision matters.

\begin{figure*}[h]
    \centering
    \includegraphics[width=0.9\textwidth]{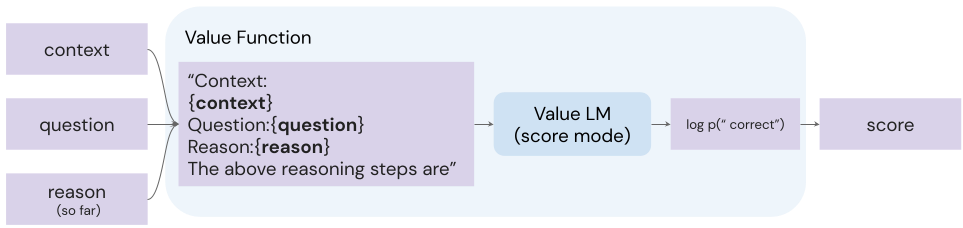}
    \caption{\textbf{The value function.} Given the context, question and a partial reasoning trace the model predicts the log probability that the current step is correct.}
    \label{fig:value_fn}
\end{figure*}

\subsection{Search}

The Value LM is trained to predict whether the current step of a reasoning trace is \textit{` correct'} or \textit{` incorrect'}. Again, we use the ground truth reasoning traces to construct examples of correct and incorrect partial reasoning traces. Constructing the correct examples is simple; we take a ground truth trace with $N$ steps and construct the following input for all $n \in [1, 2, ..., N]$, \textit{`Context:\{context\} Question:\{question\} Reason:\{reason[1:n]\} The above reasoning steps are'}. The target is \textit{` correct'} for all of these examples. To create the negative examples we take each positive example and replace one of the correct supporting statements with a different, randomly chosen statement from the context and use our Inference LM to predict the entailment. These training examples have the target \textit{` incorrect'}. Examples for both ProofWriter and EntailmentBank are shown in Figures \ref{fig:eb_search_train} and \ref{fig:pw_search_train}.

\section{Experiments and Results}

We present results on both ProofWriter (PW) \citep{tafjord2020proofwriter} and EntailmentBankQA (EB). We show that our model achieves 88.1\% and 78.1\% final answer accuracy on PW and EB respectively significantly outperforming baseline models (Table \ref{tab:main_pw_results} and \ref{tab:main_eb_results}). We also perform an ablation to demonstrate the key role of search in our model (Table \ref{tab:main_pw_results} and \ref{tab:main_eb_results}). Compared to baseline models, we show that our model often has higher reasoning trace accuracy; this is most evident on the more challenging tasks, for example PW depth-5 and EB Task 2 (Figure \ref{fig:pw_proof_eval} and \ref{fig:eb_proof_eval}). Finally, we evaluate reasoning trace validity (Section \ref{sec:trace_valid}) showing that baseline model are less likely to leverage the context when answering questions (Table \ref{tab:pw_incomplete_context} and \ref{tab:eb_random_context}) and are more likely to ``hallucinate'' statements than SI (Table \ref{tab:pw_made_up} and \ref{tab:eb_made_up}). All results in this paper were obtained using 7B parameter Chinchilla language model models \citep{hoffmann2022training}.

\subsection{Baselines}

We consider three baseline models. A \textbf{Proof + Answer} baseline where the LM is trained to predict the whole proof followed by the answer. A \textbf{Proof Only} baseline where the model is trained to predict only the proof. We use the \textbf{Proof Only} baseline to ablate the SI model by pairing it with our halter and search methods (see Tables \ref{tab:main_eb_results} and \ref{tab:main_pw_results}). Finally, we include \textbf{EntailmentWriter + Answer}. This is the entailment model of \cite{Dalvi2021ExplainingAW}, which is fine-tuned to predict an entailment tree alone, extended for question-answering by training the model to predict the answer after the final conclusion.

While \textbf{EntailmentWriter + Answer} and \textbf{Proof + Answer} tend to be very good at predicting the intermediate inferences (See Figures \ref{fig:pw_inter_iou} and \ref{fig:eb_inf_prec}) they tend to be less good at selecting the correct statements (see Figures \ref{fig:pw_leaf_iou} and \ref{fig:eb_leaf_jaccard}) and overall they perform less well on final answer accuracy (see Table \ref{tab:main_pw_results} and \ref{tab:main_eb_results}). This suggests that the models are predicting the correct intermediate outputs, without selecting the correct supporting statements and that the models are unable to use the reasoning trace to answer the question. We also see that baseline models, with the exception of EntailmentWriter, often make-up facts when reasoning (see Table \ref{tab:eb_made_up}), suggesting that their traces are not \textit{connected} and therefore are not valid (Defn. \ref{defn:connected}). Finally, baseline models leverage information in the context less well than our model, see Table \ref{tab:eb_random_context}, and Table \ref{tab:pw_incomplete_context} suggests that on ProofWriter, SI is the only model to consistently leverage the reasoning trace to answer questions.

On inspection of EntailmentWriter \citep{Dalvi2021ExplainingAW} outputs on the ProofWriter dataset we see that the model often `cheats', where the final inference helps to answer the question, but does not follow from the previously selected statements. See Section \ref{sec:EW_on_PW_cheating}. Our inference model does not have access to the question and therefore does not have the ability to cheat in this way.

\subsection{Final Answer Accuracy}

Tables \ref{tab:main_pw_results} and \ref{tab:main_eb_results} show final answer accuracy on the ProofWriter (PW) and EntailmentBankQA (EB) datasets respectively. Each table shows a comparison to baselines as well as an ablation; comparing both \textbf{SI + Halter} and the \textbf{Proof Only + Halter} baseline model with and without search.
We see that SI outperforms \textbf{EntailmentWriter + Answer} and \textbf{Proof + Answer} baseline models on all PW and EB tasks. We also show that search improves both baseline and SI performance, providing the most significant improvement for problems that require more reasoning steps (PW, depth-5) and on problems with distractors in the context (EB, Task 2)

\begin{table*}[h]
    \centering
    \begin{tabular}{l|cccc|c}
       Experiment &                      depth-1 & depth-2 & depth-3 & depth-5 & Overall\\
        \hline
        \hline
        Entailment Writer \citep{Dalvi2021ExplainingAW} + Answer         & 50.4\% & 55.3\% & 52.2\% & 56.0\% & 53.5\% \\  
        Proof + Answer                     & 70.9\% & 65.0\% & 65.5\% & 60.4\% & 65.4\% \\
        Ground truth proof + Halter        & 99.9\% &  100\% & 100\% & 100\%  & 100\% \\
        \hline
        Proof Only + Halter            & 97.0\% & 93.1\% & 84.8\% & 44.6\% & 79.9\% \\
        Proof Only + Halter + EB Search   & 99.2\% & 96.2\% & 91.4\% & 54.9\% & 85.0\% \\
        Proof Only + Halter + PW Search   & 98.7\% & 96.0\% & 90.3\% & 56.8\% & 85.4\% \\
        \hline
        SI model + Halter                  & 98.3\% & 94.1\% & 82.4\% & 38.4\% & 78.3\% \\
        SI model + Halter + EB Search         & \textbf{99.4\%} & 98.0\% & 91.7\% & 61.7\% & \textbf{88.0\%} \\ 
        SI model + Halter + PW Search & \textbf{99.4\% }& \textbf{98.1\%} & \textbf{92.0\%} & \textbf{63.4\%} & \textbf{88.1\%} \\
    \end{tabular}
    \caption{\textbf{ProofWriter ablation and comparison to baselines.} Note that the baseline model does not produce faithful reasoning traces and has access to the question when answering. By contrast, in SI the reasoning \textit{is} faithful and the answer depends on the reasoning trace. We show results using search with a value function trained on ProofWriter, \textit{PW Search}, and with a value function trained on EntailmentBank, \textit{EB Search}.}
    \label{tab:main_pw_results}
\end{table*}

On the EB dataset we see that \textbf{SI model + Halter + Search} yields similar performance to \textbf{Proof Only + Halter + Search} while also providing faithful reasoning traces, which the \textbf{Proof Only} models do not. In fact, Table \ref{tab:eb_made_up} shows that the \textbf{Proof Only} models are prone to hallucinating facts in up to 40\% of problems, while SI has made up facts to only 1\% of problems\footnote{This is likely a failure of the Selection model to produce an output with the correct syntax and could be filtered for.}. In the next section we look at reasoning trace accuracy.

\begin{table}[]
    \centering
    \begin{tabular}{l|cc}
       Model &                        Task 1 & Task 2\\
        \hline
        \hline
        Ground truth proof + Halter        &  88.8\% &  88.8\%  \\
        \hline
        Proof + Answer & 64.6\% & 7.8\%\\
        EntailmentWriter* + Answer & 50.0\% & 35.0\%  \\
        \hline
        Proof Only + Halter            & 78.5\% & 60.3\% \\
        Proof Only + Halter + Search   & 82.9\% & \textbf{76.2\%} \\
        \hline
        SI model + Halter                  & 72.4\% & 55.9\% \\
        SI model + Halter + Search          & \textbf{83.2\%} & 72.9\% \\ 
    \end{tabular}
    \caption{\textbf{EntailmentBankQA ablation and comparison to baselines.} Note that the baseline models are not causal. We use 7B parameter LMs for all models. *\citep{Dalvi2021ExplainingAW}}
    \label{tab:main_eb_results}
\end{table}

\subsection{Evaluating Reasoning Trace Accuracy}

\begin{figure*}[h]
    \centering
    \begin{subfigure}[b]{0.4\textwidth}
        \centering
        \includegraphics[width=\textwidth]{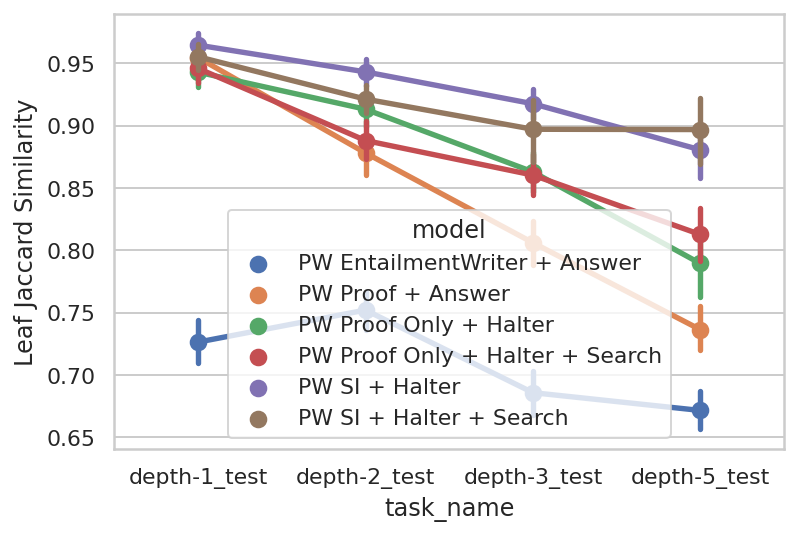}
        \caption{\textbf{Jaccard similarity between the predicted and ground-truth selection, referred to as \textit{leaves}, used to reason.} We see that the SI models perform better than Baseline models.}\label{fig:pw_leaf_iou}
    \end{subfigure}\hspace{10mm}%
    \begin{subfigure}[b]{0.4\textwidth}
        \centering
        \includegraphics[width=\textwidth]{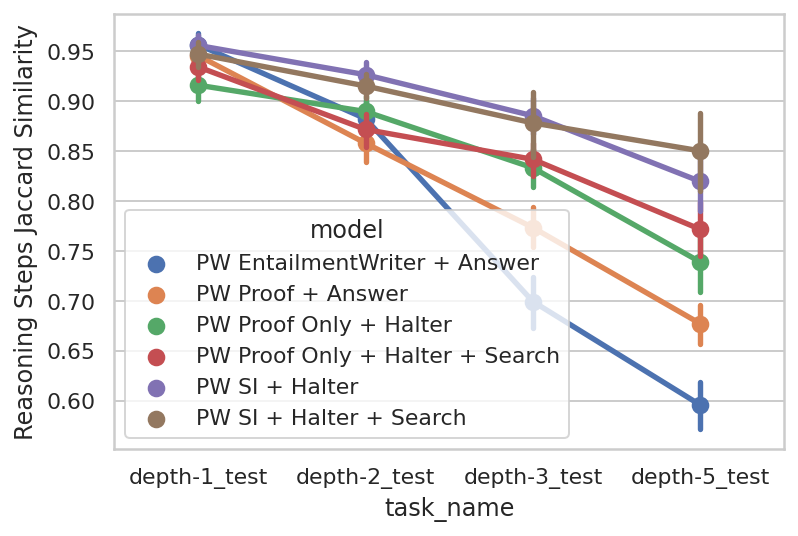}
        \caption{\textbf{Jaccard similarity between the predicted and ground-truth reasoning steps.} We see that the SI models perform better than Baseline models on the more challenging Task 2.}
        \label{fig:pw_step_iou}
    \end{subfigure}
\caption{\textbf{Evaluating proof steps for ProofWriter.} We compute the above values only on problems where the model predicts that the answer is not "Unknown".  Additional analysis in Figure \ref{fig:pw_proof_eval_extra}.}\label{fig:pw_proof_eval}
\end{figure*}

Here we evaluate the reasoning trace accuracy of each model on the PW and EB datasets, see Figures \ref{fig:pw_proof_eval}, \ref{fig:pw_proof_eval_extra} and \ref{fig:eb_proof_eval}.

Evaluating reasoning trace accuracy on PW is straightforward since we are able to use exact string match to check whether two strings are the same. We show the Jaccard similarity between predicted and ground truth leaves (i.e the selection, Figure \ref{fig:pw_leaf_iou}), intermediate outputs (i.e. the inferences, Figure \ref{fig:pw_inter_iou}) and steps (i.e. selection and inference, Figure \ref{fig:pw_step_iou}). Results show that SI had the highest Jaccard similarity for leaves and full traces while \textbf{Proof + Answer} and \textbf{Entailment Writer + Answer} have highest Jaccard similarity for intermediate outputs \ref{fig:pw_inter_iou}. This suggests that these models are correctly predicting the intermediate outputs, but not via the correct reasoning. Note that this evaluation does not consider the ordering of the proof steps which may be inflating the perceived performance of the baseline models since the baseline models are able to cheat by predicting later reasoning steps without computing earlier reasoning steps.

Overall, on the EB dataset, we see that SI outperforms \textbf{EntailmentWriter + Answer} and \textbf{Proof + Answer} baselines on the more challenging task, Task 2, which has distractors in the context. Figure \ref{fig:eb_proof_eval} shows the Jaccard similarity between predicted and ground-truth leaves (i.e. the selection) and the as well as the rouge scores between predicted and target intermediate outputs on the EB dataset (additional results in Figure \ref{fig:eb_rouge_inter_in_order}).

Note that high baseline performance on the intermediate outputs (Figures \ref{fig:pw_proof_eval_extra} and \ref{fig:eb_proof_eval}) also suggests that the baseline models have been trained well and means that their poor final answer accuracy cannot be attributed to poor training but rather to the baseline models' inability to use the reasoning trace to answer the question.

\subsection{Trace Validity}\label{sec:trace_valid}

While requirements for Defn. 1-3 are satisfied by the causal structure of our underling model (Figure \ref{fig:compare_causal}). Requirements of \textit{correctness}, for Defn. \ref{defn:valid}, are less strongly enforced. Never the less we show bellow that, unlike baseline models, our model is not able to cheat and therefore, the \textit{correctness} assumption is more likely to hold. First, however, we demonstrate that while SI satisfies the requirement of being \textit{connected}, other baseline models fail to do so.

\subsubsection{SI produces connected traces}
For a reasoning trace to be \textit{connected} it must not hallucinate facts (Defn. \ref{defn:connected}). Tables \ref{tab:eb_made_up} and \ref{tab:pw_made_up} show that while some baseline models fail to satisfy this requirement and often hallucinate facts. For example, the \textbf{Proof + Answer} baseline makes up facts to solve 60\% of EntailmentBankQA problems. On the other hand, SI makes up facts < 1\% of the time, suggesting that >99\% of traces produced by SI are connected reasoning traces.

\subsubsection{SI produces correct inferences}
Following Defn. \ref{defn:valid} for a trace to be valid it must be connected (as above) and the steps must be \textit{correct}; the inference must follow from the selection. Tables \ref{tab:pw_inference} shows that when fed with a valid selection the inference model reliably produces the \textit{correct} inference. It is harder to evaluate inference accuracy on EntailmentBankQA, however Table \ref{tab:eb_inference} suggests that the inference model is accurate, with a RougeL score if 0.69.

\subsubsection{SI uses its reasoning trace to answer the question}
Unlike baseline models, SI's causal structure (see Figure \ref{fig:compare_causal}) forces it to use the reasoning trace to answer the question. On the other hand, some baseline models are able to `cheat', answering questions without reasoning properly over the context. In other words, they depend more on the knowledge embedded in their weights than on the context provided and the reasoning trace constructed. To investigate this, we evaluate performance of a model that is given an incorrect context (a context different from the one needed to solve the problem) and compare this to performance when the model is given the correct context. If a model's answer depends on careful reasoning over the context, then it should be {\em unable} to answer the question when provided with a random context.

On the EntailmentBankQA dataset, we use a \textit{random context} sampled from another problem in the dataset. Table \ref{tab:eb_random_context} shows that both the \textbf{Proof + Answer} and \textbf{EntailmentWriter + Answer} models are still able to answer 30\% and 23\% of questions respectively, while \textbf{SI + Halter} is only able to answer 9\%. We also see that while almost half of the final accuracy could be accounted for by `cheating' or chance in the baseline models, less that 12.5\% of \textbf{SI + Halter} final accuracy could be attributed to `cheating' or chance.

On the ProofWriter dataset, we use an \textit{incomplete context} which consists only of the rules needed to solve the problems but not the facts, making it impossible to construct the correct a valid trace to solve the problem. Table \ref{tab:pw_incomplete_context} shows model performance and the difference in performance, $\Delta$, between models that use the complete and incomplete context. The $\Delta$ results suggest that \textbf{SI + Halter} is the only model that reliably makes use of the reasoning trace, while the other models rely on taking short cuts. For example, \textbf{Proof + Answer} may be taking short cuts, by looking for rules whose head predicate matches the predicate in the question.

\begin{table}[h]
    \centering
    \begin{tabular}{c|c c}
              &  \multicolumn{2}{c}{Task 1} \\
        Model & random $\downarrow$ &
              $\Delta$  $\uparrow$\\
              & context & \\
        \hline
        \hline
        SI + Halter & \textbf{9.4\%} & \textbf{63.0\%} \\
        Proof + Answer & 30.0\% & 34.6\% \\
        EW* + Answer & 23.0\% & 27.0\% \\
         
    \end{tabular}
    \caption{\textbf{EntailmentBank: Relative performance increase, $\Delta$, when using the correct context as opposed to a random one.} We expect models that actively make use of the context to have poor performance when using the \textit{random context} and a larger performance increases, $\Delta$, when using the correct context compare to when using the incorrect one. (*EW=EntailmentWriter \citep{Dalvi2021ExplainingAW})}
    \label{tab:eb_random_context}
\end{table}

\begin{table}[h]
    \centering
    \begin{tabular}{c|c c}
              &  \multicolumn{2}{c}{Depths 1-5} \\
        Model & incomplete $\downarrow$ &
              $\Delta$  $\uparrow$\\
              & context & \\
        \hline
        \hline
        SI + Halter &  \textbf{29.5\%} & \textbf{48.8\%}\\
        Proof + Answer &  61.2\% & 4.3\%\\
        EW* + Answer &  53.4\% & 0.1\% \\
         
    \end{tabular}
    \caption{\textbf{Proof Writer: Relative performance increase, $\Delta$, when using the complete context as opposed to a incomplete (rules only) context.} In the ProofWriter dataset information needed to solve the problem may be leaked, in the baseline models, by the rules themselves, without need to do valid reasoning. We expect models that actively make use of the reasoning trace -- rather than `cheating' using short-cuts -- to have \textit{poor} performance when using the \textit{incomplete context} and to have a larger performance increases, $\Delta$. (*EW=EntailmentWriter \citep{Dalvi2021ExplainingAW}) \textbf{SI + Halter} performance is less than 50\% because in 69.7\% of cases the model correctly predicts that it cannot answer the question. The $\Delta$ results suggests that \textbf{SI + Halter} is the only model that reliably uses the reasoning trace to answer questions.}
    \label{tab:pw_incomplete_context}
\end{table}

\begin{figure*}
    \centering
    \begin{subfigure}[b]{0.4\textwidth}
        \centering
        \includegraphics[width=\textwidth]{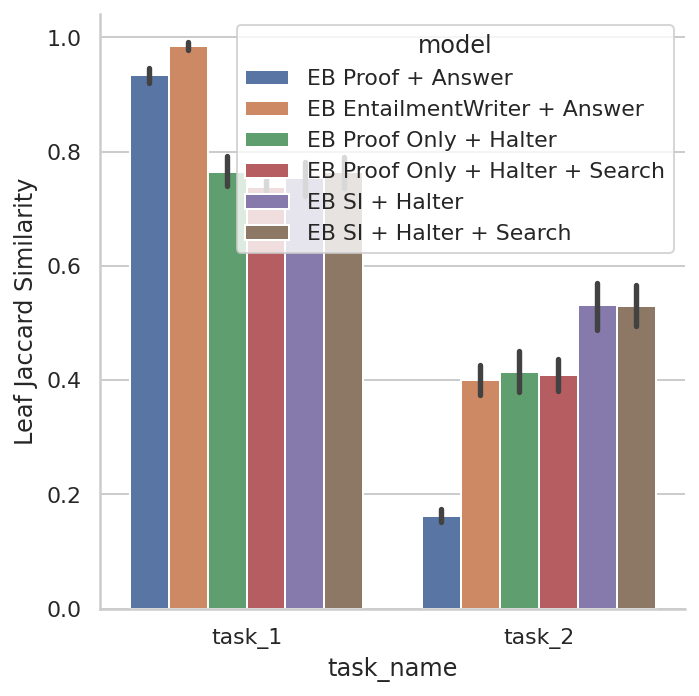}
        \caption{\textbf{Jaccard similarity between the ground truth leaves (e.g. selection) and those used by the model.} We see that SI outperforms all of the baseline models on the more challenging task, Task 2.}
        \label{fig:eb_leaf_jaccard}
    \end{subfigure}\hspace{10mm}%
    \begin{subfigure}[b]{0.4\textwidth}
        \centering
        \includegraphics[width=\textwidth]{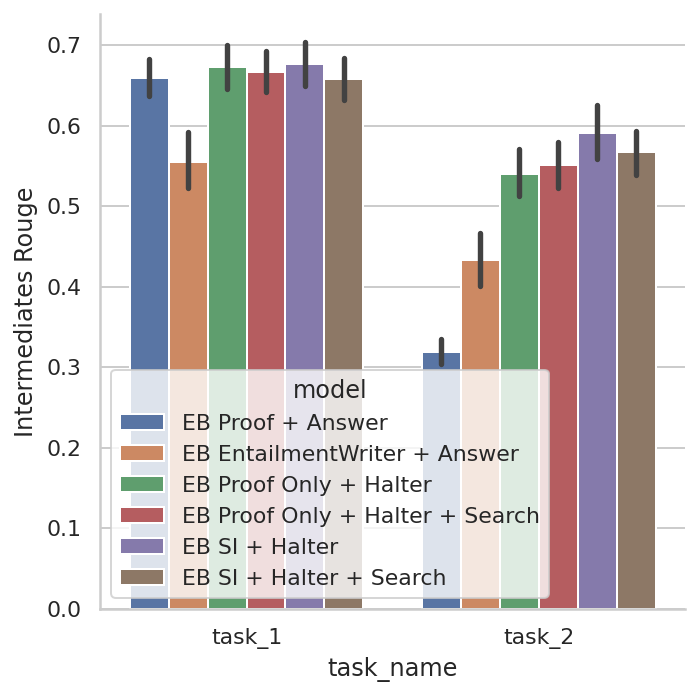}
        \caption{\textbf{Rouge score on the intermediate outputs (or inferences) from each step (ignoring order).} The baseline models that do not use search or the halter perform poorly on Task 2.}\label{fig:eb_inf_prec}
    \end{subfigure}
\caption{\textbf{Evaluating reasoning steps for EntailmentBankQA.} We compute the above values only on problems where the model predicts that the answer is not "Unknown". Note, none of these metrics account for order of the reasoning steps. }\label{fig:eb_proof_eval}
\end{figure*}

\begin{table}[h]
    \centering
    \begin{tabular}{l|c c}
    Model & Task 1 & Task 2 \\
    \hline
    \hline
    Proof + Answer &  10\% & 60\% \\
    EntailmentWriter + Answer & 3\% & \textbf{0\%} \\
    \hline
    Proof Only + Halter &  15\% & 23\% \\
    Proof Only + Halter + Search & 18\% & 40\% \\
    \hline
    SI + Halter & \textbf{1\%} & \textbf{0\%} \\
    SI + Halter + Search & \textbf{1\%} & \textbf{0\%} \\
    \end{tabular}
    \caption{\textbf{EntailmentBankQA: Proportion of problems on which models made-up facts that were not in the context.} We see that only SI and EntailmentWriter are able to avoid making up facts.}
    \label{tab:eb_made_up}
\end{table}

\section{Related Work}

While contemporary language models (LMs) are good at many natural language tasks, they often struggle with logical reasoning \citep{rae2021scaling, creswell2022selection, dasgupta2022language, betz2020critical, zhang2022paradox}. In this section we draw attention to the exciting progress being made towards reasoning using LMs. We highlight several works that use language models to produce reasoning traces \citep{wei2022chain, zelikman2022star, kojima2022large, Dalvi2021ExplainingAW, bostrom2022natural, tafjord2020proofwriter, saha2020prover}, and assess the reasoning validity of each approach. Finally, we discuss two additional areas of related work, the use of search and the problem of when to stop reasoning.

\begin{figure*}[h]
    \centering
    \includegraphics[width=\textwidth]{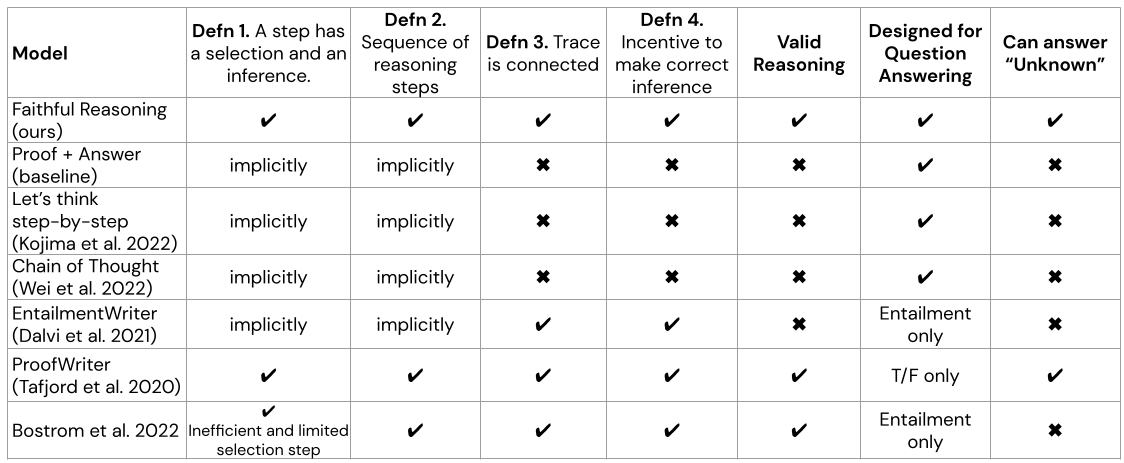}
    \caption{\textbf{Comparison between Faithful Reasoning and other related works.}}
    \label{fig:compare_table}
\end{figure*}

\subsection{Language Models Are Not Enough}

Recent work on applying language models to reasoning problems has largely concentrated on improving final answer accuracy rather than producing valid, human interpretable reasoning traces that lead to the answers. For example, various methods of prompting \citep{wei2022chain} and iterative fine-tuning \citep{zelikman2022star} have been used to encourage models to produce reasoning traces, and while this has led to improvement of final answer accuracy these traces do not support our understanding of how the answer was reached.

\cite{kojima2022large} split the reasoning in two parts, first producing a reasoning trace and then predicting the answer given the question and the reason. Similar to our own model \cite{zhou2022least} go one-step further and split each reasoning step in two: first asking an intermediate question and second, answering that question. While the authors suggest that their approach promotes compositional generalisation, unlike our approach the answering part of the model has full access to the question and therefore the model does not have to rely on the reasoning trace to answer the question. Moreover, unlike our work, the models of \cite{zhou2022least, kojima2022large, zelikman2022star, wei2022chain} are not restricted to reasoning over knowledge in the context, but rather have the ability to hallucinate possibly incorrect ``knowledge'' to support the answer leading to reasoning traces which are not valid and cannot be trusted.

\subsection{Reasoning with Language Models}

The EntailmentBank dataset proposed by \cite{Dalvi2021ExplainingAW} has led to several works focused on deriving reasoning traces to backup an answer or hypothesis \citep{ribeiro2022entailment, dalvi2022towards, jhamtani2020learning, bostrom2022natural}. In our work, we focus on answering questions and providing faithful reasoning traces, rather than post-hoc explanations. 

With a similar motivation to our own, \cite{gupta2022right} and \cite{nakano2021webgpt} show promising results extracting evidence from a table or the web, respectively, and using this to answer a question or solve a natural language inference (NLI) problem. However, while \cite{gupta2022right} and \cite{nakano2021webgpt} show the evidence used, they do not show how that information was combined to answer the question. In our work, we produce a valid reasoning trace that shows how multiple pieces of knowledge are combined, over several iterations, to answer a question.

Other works have focused on using reasoning to show whether a statement is True or False \citep{tafjord2020proofwriter, betz2020critical}. In ProofWriter, \cite{tafjord2020proofwriter} train an LM to enumerate implications (and corresponding reasoning steps) given a hypothesis. A valid reasoning trace can be constructed from these outputs. However, this approach is limited to answering questions whose answer is True, False or Unknown, and a reasoning trace must be constructed post-hoc.

Finally, while several works have informally introduced the notion of faithful reasoning \citep{kumar2020nile, gupta2022right, bostrom2022natural}, we have related this more precisely to the definition of \textit{valid} reasoning in logic.

\subsection{Using Search for Reasoning Problems}

The notion of valid and invalid reasoning traces have also been explored in the context of search. \cite{jhamtani2020learning} develop datasets of valid and invalid reasoning traces for grade school science questions. These can be used to train models to detect valid reasoning traces. However, it can be expensive to collect both valid and \textit{invalid} reasoning traces hence they collect only shallow traces and their traces do not include intermediate inferences. Instead, we show how, given a valid reasoning trace, we can generate many invalid reasoning traces that can be used to fine-tune a value function and used to guide search. Also, rather than learning a verifier that evaluates a whole trace \citep{jhamtani2020learning, nye2021show, cobbe2021training}, we train a model on partial reasoning traces, resulting in a model more similar to a value function which assesses the ``value'' of the current reasoning step, which can be used for step-level search.

\cite{bostrom2022natural} also use step-level search to determine whether a hypothesis is entailed by a set of statements. While we perform a beam search, using a learned value function, to find high-quality reasoning traces, \cite{bostrom2022natural} depend on exhaustive search to evaluate all possible pairs of statements to use for selection. Unlike \cite{bostrom2022natural}, our selection step is not limited to selecting just two statements. This allows us to more efficiently solve ProofWriter tasks whose rules may be conditioned on multiple statements.

\subsection{The Problem of When to Stop Reasoning}

The problem of when to ``stop'' rarely features in the deep learning literature because our models typically answer problems in a single step. However, there are some exceptions. A simple example is text synthesis with large language models where the model has to determine when to stop producing tokens. This is often handled by a special \textit{`End Of Sequence'} token \citep{graves2013generating}. Other examples in the deep learning literature draw random variables from a parameterised distribution to predict when to stop reasoning \citep{banino2021pondernet, graves2016adaptive}.

Related work by \cite{kadavath2022language} also investigates when LMs ``know'' the answer. Their model proposes a number of candidates, and predicts whether each candidate is the answer to the question or not. Additionally, \cite{bostrom2022natural} tackle the less challenging problem of determining whether an inference matches a goal state.

In summary, current work focuses on True/False/NLI tasks \citep{bostrom2022natural, Dalvi2021ExplainingAW, tafjord2020proofwriter} while our work tackles question-answering. This is not a trivial difference. In question-answering, there is less information with which to construct the reasoning trace, since the ``goal'' is not known, and learning when to terminate is also more challenging. Moreover, current work leverages reasoning traces to boost performance -- rather than to aid explainability or build trust -- allowing for hallucination of ``knowledge'' during reasoning \citep{wei2022chain, kojima2022large}. Furthermore, some existing approaches still allow the opportunity for ``cheating'' \cite{Dalvi2021ExplainingAW, wei2022chain} by providing the answering part of the model with direct access to the question\footnote{Specifically in these cases, the question itself contains sufficient information to supply the answer, unlike in ProofWriter where the question is also not sufficient for answering correctly.}. Finally, unlike most other models \citep{wei2022chain, Dalvi2021ExplainingAW}, the causal structure of our model (see Figure \ref{fig:compare_table}) mirrors the requirements for validity, see Table \ref{fig:compare_table}. Other approaches that do satisfy validity have their own limitations, as detailed above.

\section{Limitations}

The causal structure of our model mirrors the requirements for producing a {\em valid} trace (Defn. \ref{defn:valid}). Requirements for a {\em connected} reasoning trace (Defn. \ref{defn:connected}) are guaranteed by design (Section \ref{sec:selection}). Unavoidably, given our use of LMs, we cannot guarantee that all reasoning steps will be logically correct (Defn. \ref{defn:valid}). However, our architecture is designed to encourage logical correctness by preventing models from `cheating'. For example, if the Selection model selects two unrelated statements, then the Inference model may draw a nonsensical conclusion. We also mitigate this by introducing a learned value function (Section \ref{sec:search}) that filters out poor reasoning traces, although this still cannot guarantee the correctness of every step. Examples of both correct and incorrect reasoning traces, along with their value (according to the value function), are shown in Section \ref{sec:traces}.

In this paper we have focused on developing models that answer questions using valid reasoning. For now we have assumed access to a context, over which to reason. However, while there are some settings where such a context may be provided, in most real world settings this is unlikely. In this paper we have chosen to focus on the challenging problem of multi-step reasoning within a {\em given} context. However, in future work we hope to incorporate retrieval to populate the context, and there is already interesting research in this direction \citep{ribeiro2022entailment, Dalvi2021ExplainingAW, xie2020worldtree}.

\section{Discussion}

Language models are being applied, with great success, to many different problems \cite{zeng2022socratic, rae2021scaling, nakano2021webgpt, alayrac2022flamingo, nye2021show}. However, they largely remain black boxes; we do not know how the models produce their responses. One solution to this is to develop models that can produce faithful reasoning traces. We characterise faithful reasoning in terms of logical validity (Section \ref{sec:defn}), and propose Selection-Inference, a model that mirrors the structure of this definition, and is guaranteed to produce valid reasoning traces under the assumption that individual steps are correct (Defn. \ref{defn:valid}). By fine-tuning an Inference model specifically for this task and preventing it from ``cheating'', we increase the likelihood that this assumption holds (Tables \ref{tab:pw_inference}and  \ref{tab:eb_inference}). Finally, to find high-quality reasoning traces, we introduce a value function, and use it to guide a beam search through the tree of potential traces induced by the non-determinism of selection.

The resulting model achieves higher final answer accuracy than baseline models on both ProofWriter \cite{tafjord2020proofwriter} and EntailmentBankQA \cite{Dalvi2021ExplainingAW} tasks. We see that both \textbf{Proof Only} and \textbf{SI} benefit from search (Tables \ref{tab:main_pw_results} and \ref{tab:main_eb_results}). When compared to baseline models, our model is less likely to hallucinate facts while reasoning (Table \ref{tab:pw_made_up} and \ref{tab:eb_made_up}). We see that the \textbf{SI + Halter} model is far more likely than baseline models to pay attention to the context (Table \ref{tab:eb_random_context}) and to leverage the reasoning trace (Table \ref{tab:pw_incomplete_context}). Overall, we see that \textbf{SI + Halter (+ Search)} models achieves superior reasoning trace accuracy, especially on the more challenging tasks (Figures \ref{fig:pw_proof_eval} and \ref{fig:eb_proof_eval}).

Our approach exemplifies a trend towards {\em algorithmic prompting}, a form of automated prompt engineering in which querying a language model becomes a computational primitive. The responses of the language model can be manipulated to construct new prompts that are then used to make further queries. Model queries and prompt construction are composed into algorithms with the usual computational constructs: sequence, choice, and iteration. Algorithmic prompting can be used to elicit more sophisticated and nuanced behaviour from a language model than would otherwise be possible. For example, as our work shows, this approach can be used to develop models capable of faithful reasoning, without compromising performance. In future work we aim to leverage advancements in retrieval to populate the context, rather than relying on the context being provided in the question.

\bibliographystyle{abbrvnat}
\nobibliography*
\bibliography{main}

\section*{Acknowledgements}
The authors would like to thank Angeliki Lazaridou, Charles Blundell and Christopher Summerfield for feedback on our paper as well as Jonathan Uesato, Jordi Grau-Moya, Ramana Kumar and Irina Higgins for insightful discussions.

\pagebreak
\appendix
\section*{Supplementary Materials}

\section{Formal definition of the problem}\label{sec:formal}

Formally, suppose we have a problem of the form $(q, \mathcal{C}_0)$, where $\mathcal{C}_0$ is a context, consisting of a set of statements which are sufficient to predict the the correct answer to the question, $q$.

The role of the Selection model, $\text{LM}_{\text{Selection}}$ is to sample a selection, $s_k$, given the question, $q$ and the current context, $\mathcal{C}_{k}$, see Equation \ref{eqn:selection}.

\begin{equation}\label{eqn:selection}
    s_k  = \text{LM}_{\text{Selection}}(q, \mathcal{C}_{k})
\end{equation}

The role of the Inference model,  $\text{LM}_{\text{Inference}}$ is to sample an inference, $i_k$, given the selection, $s_k$, see Equation \ref{eqn:inference}.

\begin{equation}\label{eqn:inference}
    i_k = \text{LM}_{\text{Inference}}(s_k)
\end{equation}

After each inference the context is updated as follows, $\mathcal{C}_{k-1} \cup i_{k-1}$, accumulating inferences from previous steps of reasoning.

The Halter LM, $\text{LM}_{\text{halt}}$, is applied to each inference, $i_t$, in two ways. First, to choose whether the model should stop reasoning and the second to answer the question when the model is `ready'. This is illustrated in Alg. \ref{alg:si} and Alg. \ref{alg:halt}.

\begin{algorithm}[h]
  \caption{\textbf{The SI() function}.}\label{alg:si}
  \KwIn{$\text{LM}_{\text{Selection}}$: Selection LM.}
  \KwIn{$\text{LM}_{\text{Inference}}$: Inference LM.}
  \KwIn{\textbf{halt()}: Halt function (Alg. \ref{alg:halt}).}
  \KwIn{$q$: Question.}
  \KwIn{$\mathcal{C}_0$: Initial context.}
  \KwIn{$c$: Choices.}
  \KwIn{$K'$: Max. reasoning steps.}
  $a \gets \textit{`Unknown'}$; Initial answer is unknown. \\
  $t \gets$ 0; Step counter. \\
  \While{$a$ \textbf{is} \textit{`Unknown'}}{
  $s_k \gets \text{LM}_{\text{Selection}}(q, \mathcal{C}_k)$\;
  $i_k \gets \text{LM}_{\text{Inference}}(i)$\;
  $C_{k+1} \gets \mathcal{C}_{k} \cup i_{k}$\;
  $a \gets \text{\textbf{halt}}(q, i_k, c)$\;
  $k \gets k + 1$\;
  \If{$k>K'$}{\Return $a$}
  }
  
  $ a\gets \text{LM}_{\text{halt}}(i, c)$\;
  \Return $a$\;
\end{algorithm}

\begin{algorithm}[h]
  \caption{\textbf{The halt() function.} Note that we use the same language model $\text{LM}_{\text{halt}}$ to both determine whether the model is able to answer the question and to answer the question. The key difference is the prompt, shown in Section \ref{sec:halt}.}\label{alg:halt}
  \KwIn{$\text{LM}_{\text{halt}}$: Halting LM.}
  \KwIn{$q$: Question.}
  \KwIn{$i$: Current inference.}
  \KwIn{$c$: Choices.}
  $a \gets \text{LM}_{\text{halt}}(q, i)$\;
  \If{$a$ \textbf{is} \textit{`Unknown'}}{
  \Return $a$\;
  }
  \Else{
  
  $ a\gets \text{LM}_{\text{halt}}(i, c)$\;
  \Return $a$\;
  }
\end{algorithm}

\section{Training Details}

\subsection{Datasets}
\subsubsection{ProofWriter}
\label{sec:proofwriter_sup}
The ProofWriter dataset \citep{tafjord2020proofwriter} contains both a Closes and Open World Assumption version (CWA and OWA respectively). We use a subset of the OWA dataset. This is because for the CWA dataset everything that cannot be proven is considered False. This means that problems whose answer is False, do not have reasoning traces. On the other hand the OWA dataset contains proofs for problems whose answers are True and False. Those without proofs are "Unknown". Since we need proofs for training and evaluation, we use the problems from the OWA dataset that have accompanying proofs (i.e. those whose answer is not Unknown).

\subsection{Selection-Inference}
\begin{figure}[h!]
\begin{subfigure}{\columnwidth}
    \centering
    \includegraphics[width=\textwidth]{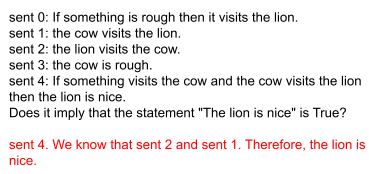}
    \caption{\textbf{Example of $\langle$input, {\color{red}target}$\rangle$ pairs used to train the Selection LLM.}}
    \label{fig:pw_select_train}
\end{subfigure}
\begin{subfigure}{\columnwidth}
    \centering
    \includegraphics[width=\textwidth]{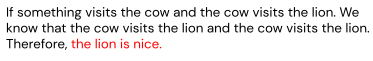}
    \caption{\textbf{Example of $\langle$input, {\color{red}target}$\rangle$ pairs used to train the Inference LLM.}}
    \label{fig:pw_infer_train}
\end{subfigure}
\caption{\textbf{Examples of ProofWriter training pairs for Selection and Inference LLMs}}\label{fig:pw_si_train}
\end{figure}

\begin{figure}[h!]
\begin{subfigure}{\columnwidth}
    \centering
    \includegraphics[width=\textwidth]{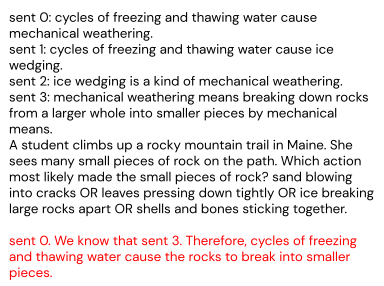}
    \caption{\textbf{Example of $\langle$input, {\color{red}target}$\rangle$ pairs used to train the Selection LLM.}}
    \label{fig:eb_select_train}
\end{subfigure}
\begin{subfigure}{\columnwidth}
    \centering
    \includegraphics[width=\textwidth]{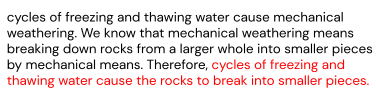}
    \caption{\textbf{Example of $\langle$input, {\color{red}target}$\rangle$ pairs used to train the Inference LLM.}}
    \label{fig:eb_infer_train}
\end{subfigure}
\caption{\textbf{Examples of EntailmentBankQA training pairs for Selection and Inference LLMs}}\label{fig:eb_si_train}
\end{figure}

Figures \ref{fig:pw_si_train} and \ref{fig:eb_si_train} show examples of training samples use to fine-tune the Selection and Inference LMs.

\subsection{Halter}

\begin{figure*}[h!]
    \centering
    \includegraphics[width=\textwidth]{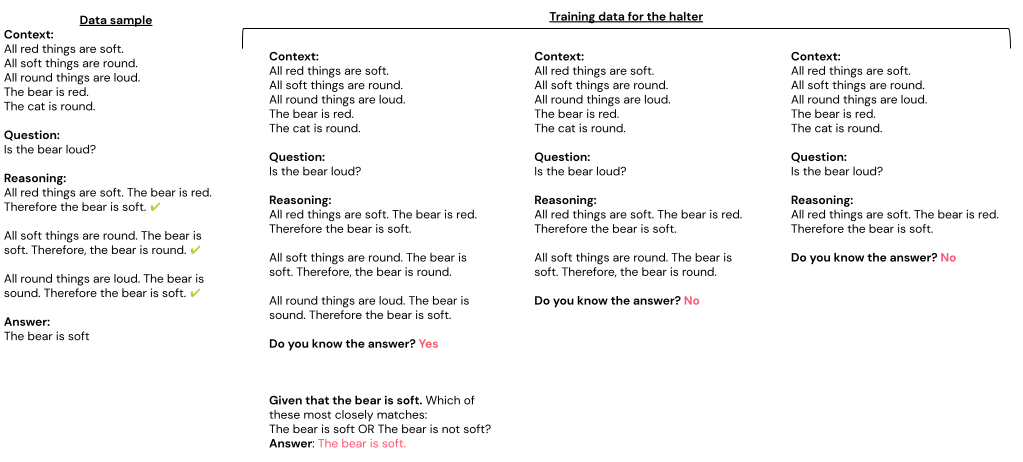}
    \caption{\textbf{Example of how data is generated for the halter.} Above are examples for four training data points. The first three, \textbf{Do you know the answer?} examples, are used to train the halting model to learn when to halt. The final datum is used to train the halter to select an answer from the available choices.}
    \label{fig:halter_train}
\end{figure*}

Figure \ref{fig:halter_train} shows how training data is constructed for training the halting model.

\subsection{Search}

\begin{figure*}[h]
    \centering
    \includegraphics[width=\textwidth]{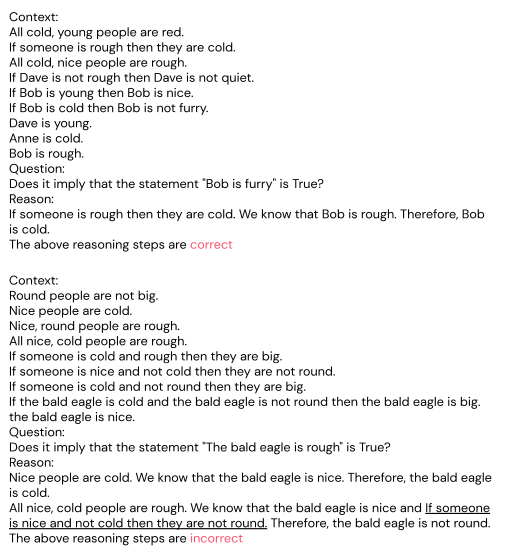}
    \caption{\textbf{Examples of correct (top) and incorrect (bottom) training data samples used to train the Value LM.} Targets are shown in {\color{red}red}. The underlined statement in the last line of the incorrect reasoning trace is the one that is substituted in. The Inference LM is used to compute the inference. It is very easy to see here that the second statement is incorrect because it contains a rule rather than a fact.}
    \label{fig:pw_search_train}
\end{figure*}

\begin{figure*}[h]
    \centering
    \includegraphics[width=\textwidth]{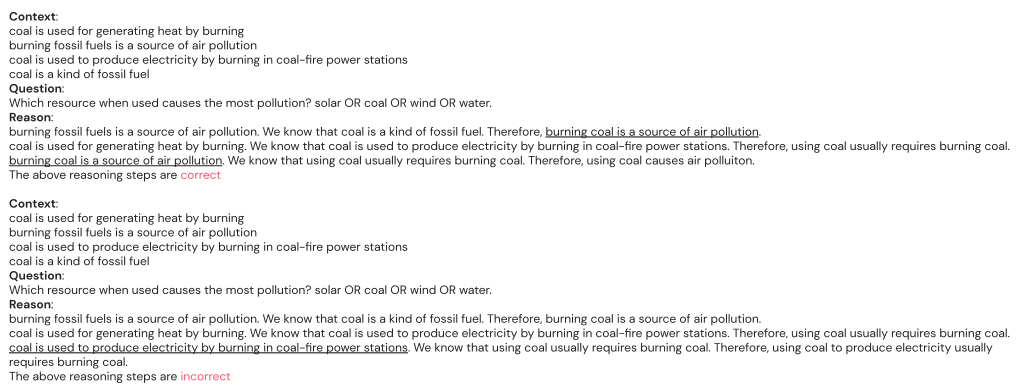}
    \caption{\textbf{Examples of correct (top) and incorrect (bottom) training data samples used to train the Value LM.} Targets are shown in {\color{red} red}. The underlined statement in the last line of the correct reasoning trace is replaced with a random, incorrect statement from the context. The Inference LM is used to compute the inference.}
    \label{fig:eb_search_train}
\end{figure*}

Figures \ref{fig:pw_search_train} and \ref{fig:eb_search_train} show examples of data points used to train the value function. The targets for the value function are either \textit{` correct'} or \textit{` incorrect'}.

\section{Additional Results}

\subsection{Halter}

\begin{figure*}[h]
    \centering
    \includegraphics[width=\textwidth]{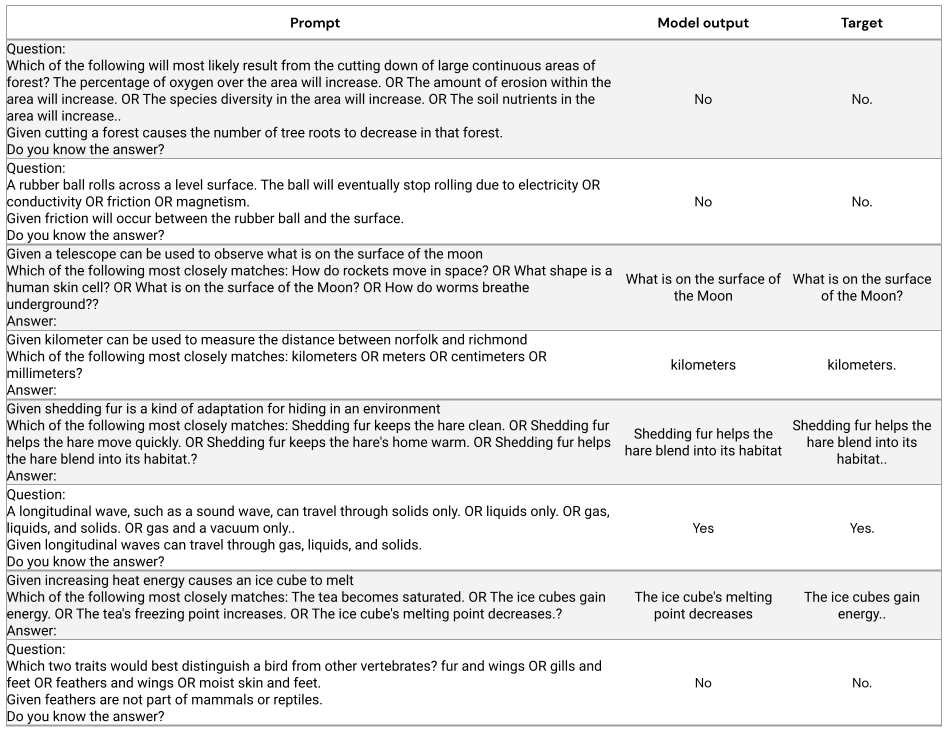}
    \caption{\textbf{Qualitative results on EntailmentBankQA showing halter outputs on the EntailmentBankQA dataset.}}
    \label{fig:halter_qual}
\end{figure*}

Figure \ref{fig:halter_qual} shows qualitative results from the Halter model trained on EntailmentBankQA.

\subsection{Reasoning Trace Accuracy}

\begin{figure*}[h]
    \begin{subfigure}{\columnwidth}
        \centering
        \includegraphics[width=\columnwidth]{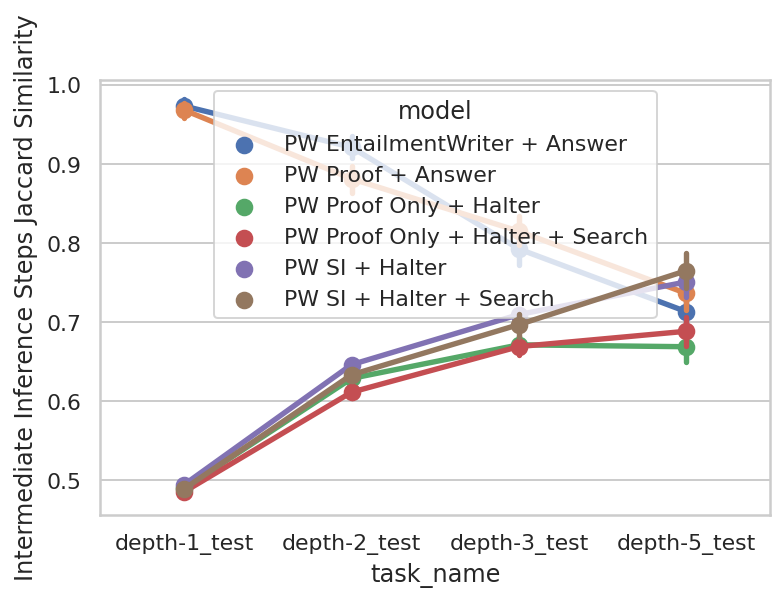}
        \caption{\textbf{Jaccard similarity between the GT intermediate steps and the predicted intermediate steps.}}
        \label{fig:pw_inter_iou}
    \end{subfigure}
    \begin{subfigure}{\columnwidth}
        \centering
        \includegraphics[width=\columnwidth]{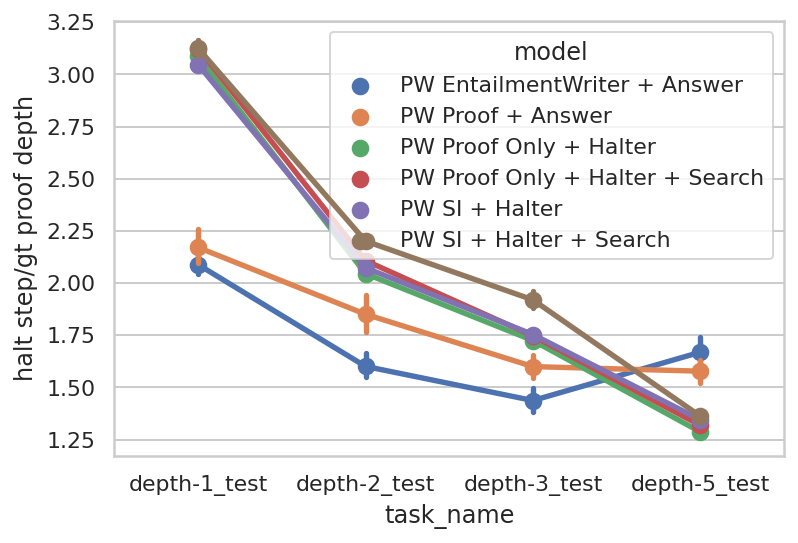}
        \caption{\textbf{Comparing the number of reasoning steps in the GT proof to those in the predicted proof.} }
        \label{fig:pw_halt_eval}
    \end{subfigure}
    \begin{subfigure}{\columnwidth}
        \centering
        \includegraphics[width=\columnwidth]{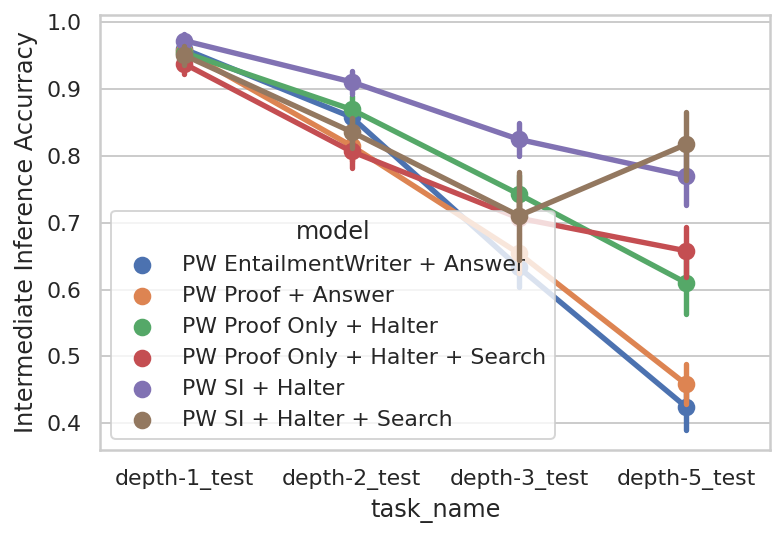}
        \caption{\textbf{Intermediate inference accuracy where order of the inferences matters.}}
        \label{fig:pw_inter_order_eval}
    \end{subfigure}
    \begin{subfigure}{\columnwidth}
        \centering
        \includegraphics[width=\columnwidth]{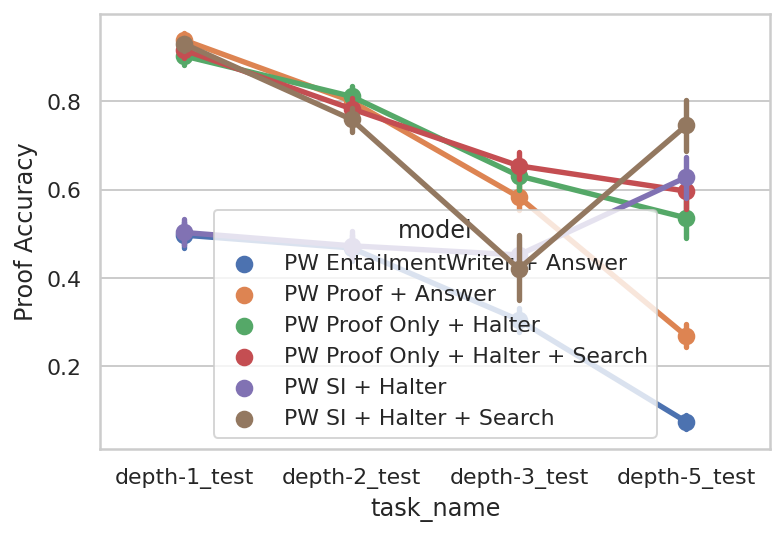}
        \caption{\textbf{Exact string match between the ground truth proof and the predicted proof.} }
        \label{fig:pw_exaxt_str_match_eval}
    \end{subfigure}
    \caption{\textbf{Evaluating reasoning traces on ProofWriter.} For exact string match we remove all removing non-alphabetic characters and compare characters in lower case.}\label{fig:pw_proof_eval_extra}
\end{figure*}

Figure \ref{fig:pw_proof_eval_extra} show additional evaluation of reasoning traces on the ProofWriter dataset.

In Table \ref{tab:eb_made_up} we saw that baseline models, with the exception of EntailmentWriter, were more likely to hallucinate facts while reasoning on the EntailmentBank dataset than SI. Interestingly, Table \ref{tab:pw_made_up} shows that \textbf{Proof + Answer} and \textbf{Proof Only} baseline models have \textit{learned} not to make up facts while reasoning on the ProofWriter dataset. Note that both EntailmentWriter and SI (ours) are \textit{designed} not to make up facts.
\begin{table*}[h]
    \centering
    \begin{tabular}{l|c c c c}
    Model & \multicolumn{4}{c}{depth} \\
     & depht-1 & depth-2 & depth-3 & depth-5 \\
    \hline
    \hline
    Proof + Answer &  0\% & 0\% & 1\% & 1\%\\
    EntailmentWriter + Answer &  0\% & 0\% & 0\% & 0\%\\
    \hline
    Proof Only + Halter &  4\% & 1\% & 1\% & 0\%\\
    Proof Only + Halter + Search &   0\% & 0\% & 0\% & 0\%\\
    \hline
    SI + Halter &  0\% & 0\% & 0\% & 0\%\\
    SI + Halter + Search &   0\% & 0\% & 0\% & 0\%\\
    \end{tabular}
    \caption{\textbf{ProofWriter: Proportion of problems on which models made-up facts that were not in the context.} We see that the \textbf{Proof + Answer} and \textbf{Proof Only} baseline models have learned not to make up facts, while EntailmentWriter and SI are designed not to make up facts.}
    \label{tab:pw_made_up}
\end{table*}

\begin{figure}[h]
    \centering
    \includegraphics[width=\columnwidth]{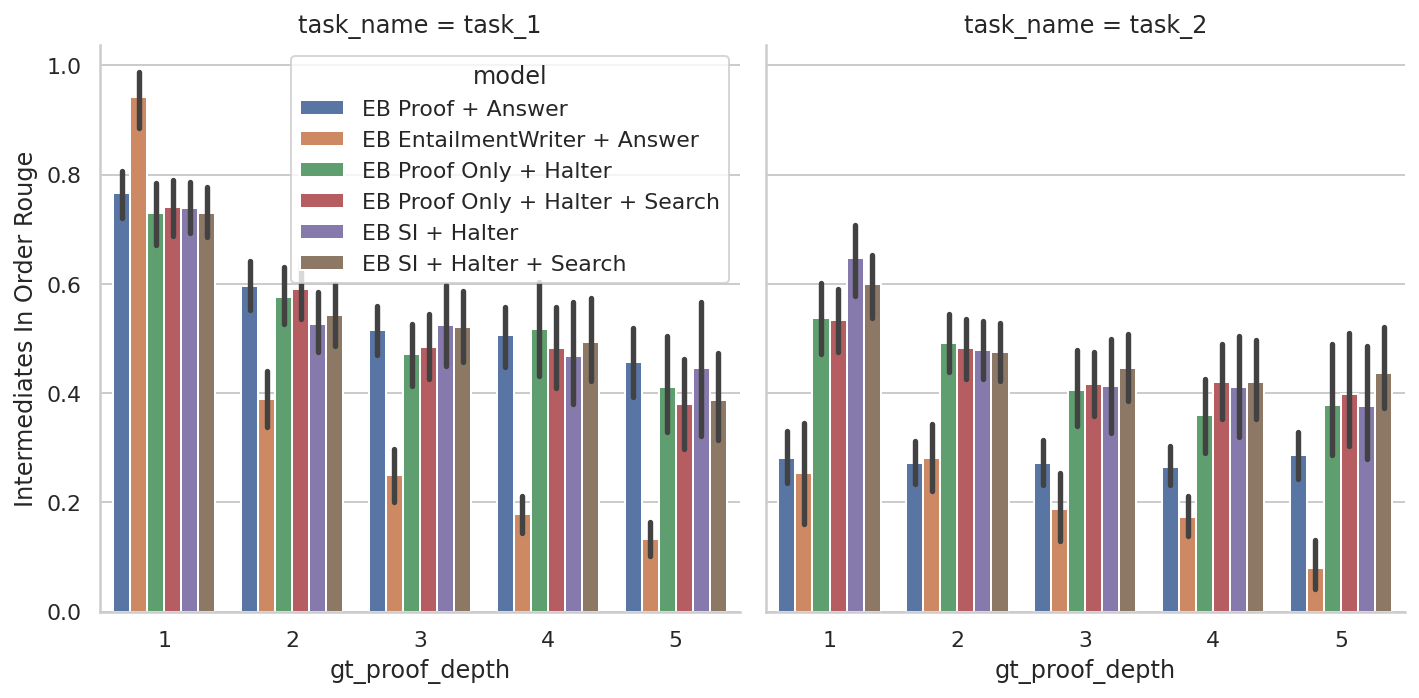}
    \caption{\textbf{EntailmentBankQA: Rouge Score between ground truth and predicted intermediate inferences in order.}}
    \label{fig:eb_rouge_inter_in_order}
\end{figure}

Figure \ref{fig:eb_rouge_inter_in_order} shows the Rouge 1 scores between the predicted and ground truth, ordered, intermediate inferences. We see that EntailmentWriter is very good at single step inference on Task 1 problems, but performance quickly declines for problems requiring multiple steps of reasoning. In general SI models and models using halting and search outperform the baseline models.

Tables \ref{tab:pw_inference} and \ref{tab:eb_inference} show the accuracy of the Inference LM when fed valid selections.
\begin{table}[h]
    \centering
    \begin{tabular}{c|c}
        Task & Inference Accuracy \\
        \hline
        depth-1 & 100\% \\ 
        depth-2 & 100\% \\ 
        depth-3 & 100\% \\ 
        depth-5 & 99.9\% \\ 
    \end{tabular}
    \caption{\textbf{ProofWriter inference accuracy.} The inference model achieves almost perfect performance. We use exact string match in lower case to decide if two statement are the same.}
    \label{tab:pw_inference}
\end{table}

\begin{table}[h]
    \centering
    \begin{tabular}{c|c}
        Metric & Inference Accuracy \\
        \hline
        Rouge1 & 0.69\\ 
        Rouge2 & 0.55\\ 
        RougeL & 0.69\\ 
        BLEURT > 0.28 & 64\% \\ 
    \end{tabular}
    \caption{\textbf{EntailmentBankQA inference accuracy.} We report Rouge scores as well as accuracy using a BLEURT \citep{sellam2020bleurt} with a threshold of 0.28 determine accuracy (as done by \cite{Dalvi2021ExplainingAW}.}
    \label{tab:eb_inference}
\end{table}

\section{Selection-Inference Model Outputs} \label{sec:traces}

\subsection{ProofWriter: SI + Halter + Search}

Below we show reasoning traces from the \textbf{SI + Halter + Search} model with the top 10 value function scores. No additional filtering is performed. For ease of reading we have combined each selection and inference into a single line of text rather than showing them separately. Examples that score high often involve repeated steps, this is because the ProofWriter proof dataset often includes repeated steps. Invalid reasoning is shown in red.

\subsubsection*{Example 1 (value: -6.7e-06)} 

\textbf{Context:} \\
If something likes the rabbit and it sees the bald eagle then the bald eagle needs the rabbit. \\
If something is blue then it sees the rabbit. \\
If something needs the rabbit and it likes the bald eagle then the bald eagle is blue. \\ 
If something likes the mouse then it is round. \\ 
If something is rough and it does not see the mouse then it does not need the bald eagle. \\ 
If something sees the mouse then it needs the rabbit. \\ 
the rabbit likes the mouse. \\ 
the bald eagle likes the rabbit. \\ 
the rabbit likes the bald eagle. \\ 
the rabbit sees the bald eagle. \\ 
the rabbit is blue. \\ 
the rabbit needs the bald eagle. \\ 
the bald eagle needs the mouse. \\ 
the rabbit needs the mouse. \\ 
the mouse needs the bald eagle. \\ 
the rabbit is not green. \\ 
the bald eagle likes the mouse. \\ 
the rabbit sees the mouse. \\ 
the bald eagle is round. \\ 
the bald eagle does not see the mouse. \\
the rabbit is red. \\
the rabbit is rough. \\
\textbf{Question:} Does it imply that the statement "The bald eagle does not see the rabbit" is True? \\
\textbf{Proof:} \\
If something sees the mouse then it needs the rabbit. We know that the rabbit sees the mouse. Therefore,  the rabbit needs the rabbit. \\
If something sees the mouse then it needs the rabbit. We know that the rabbit sees the mouse. Therefore,  the rabbit needs the rabbit. \\
If something sees the mouse then it needs the rabbit. We know that the rabbit sees the mouse. Therefore,  the rabbit needs the rabbit. \\
If something needs the rabbit and it likes the bald eagle then the bald eagle is blue. We know that the rabbit needs the rabbit and the rabbit likes the bald eagle. Therefore,  the bald eagle is blue. \\
If something is blue then it sees the rabbit. We know that the bald eagle is blue. Therefore,  the bald eagle sees the rabbit. \\
\textbf{Answer:}  False \\
\textbf{Target Proof:} \\
If something sees the mouse then it needs the rabbit. We know that the rabbit sees the mouse. Therefore, the rabbit needs the rabbit. \\
If something needs the rabbit and it likes the bald eagle then the bald eagle is blue. We know that the rabbit needs the rabbit and the rabbit likes the bald eagle. Therefore, the bald eagle is blue. \\
If something is blue then it sees the rabbit. We know that the bald eagle is blue. Therefore, the bald eagle sees the rabbit. \\
\textbf{Target Answer:} False \\

\subsubsection*{Example 2 (value: -8.6e-06)}

\textbf{Context:} \\
If someone needs the bald eagle and the bald eagle needs the squirrel then the bald eagle likes the cow. \\
If someone needs the squirrel then they are green. \\ 
If someone needs the cow and they need the dog then the cow is green. \\ 
If someone likes the squirrel and they like the dog then the dog is nice. \\ 
If someone sees the cow then they need the dog. \\ 
If someone is cold then they see the cow. \\ 
If someone needs the dog then the dog sees the cow. \\ 
If someone likes the dog and the dog is blue then the dog sees the bald eagle. \\ 
the dog is nice. \\ 
the cow is nice. \\ 
the dog likes the squirrel. \\ 
the squirrel likes the dog. \\
the dog likes the cow.  \\
the squirrel sees the dog. \\
the bald eagle is cold. \\
the bald eagle sees the dog. \\
the squirrel sees the bald eagle. \\
the dog sees the bald eagle. \\
the squirrel needs the cow. \\
the squirrel is cold. \\
\textbf{Question:} Does it imply that the statement "The cow is not green" is True? \\
\textbf{Proof:} \\
If someone is cold then they see the cow. We know that the squirrel is cold. Therefore,  the squirrel sees the cow. \\
If someone is cold then they see the cow. We know that the bald eagle is cold. Therefore,  the bald eagle sees the cow. \\
If someone sees the cow then they need the dog. We know that the squirrel sees the cow. Therefore,  the squirrel needs the dog. \\
If someone needs the cow and they need the dog then the cow is green. We know that the squirrel needs the cow and the squirrel needs the dog. Therefore,  the cow is green. \\
\textbf{Answer:} False \\
\textbf{Target Proof:} \\
If someone is cold then they see the cow. We know that the squirrel is cold. Therefore, the squirrel sees the cow. \\
If someone sees the cow then they need the dog. We know that the squirrel sees the cow. Therefore, the squirrel needs the dog. \\
If someone needs the cow and they need the dog then the cow is green. We know that the squirrel needs the cow and the squirrel needs the dog. Therefore, the cow is green. \\
\textbf{Target Answer:} False \\

\subsubsection*{Example 3 (value: -9.5e-06)}

\textbf{Context:} \\
If something is red and it likes the tiger then it needs the dog. \\ 
If the cat likes the tiger and the cat needs the dog then the dog is not cold. \\ 
If something eats the dog and the dog needs the bald eagle then the dog is nice. \\ 
If something eats the cat and it needs the dog then the cat is cold.\\ 
If the tiger needs the dog and the dog is green then the dog needs the tiger. \\
If something eats the cat then the cat needs the bald eagle. \\
All cold things are nice. \\
If the bald eagle needs the dog and the dog eats the bald eagle then the dog is not cold. \\
If something needs the bald eagle then it eats the cat. \\
the tiger needs the bald eagle. \\
the bald eagle is nice. \\
the bald eagle is round. \\ 
the cat likes the bald eagle. \\ 
the bald eagle does not eat the cat. \\ 
the cat likes the tiger. \\ 
the cat needs the tiger. \\ 
the dog likes the bald eagle. \\ 
the tiger eats the dog. \\ 
the cat eats the tiger. \\ 
the bald eagle is green. \\ 
the bald eagle likes the tiger. \\ 
the cat is red. \\ 
the dog is not round. \\
\textbf{Question:} Does it imply that the statement "The cat is nice" is True? \\
\textbf{Proof:} \\
If something is red and it likes the tiger then it needs the dog. We know that the cat is red and the cat likes the tiger. Therefore,  the cat needs the dog. \\ 
If something needs the bald eagle then it eats the cat. We know that the tiger needs the bald eagle. Therefore,  the tiger eats the cat. \\ 
If something eats the cat then the cat needs the bald eagle. We know that the tiger eats the cat. Therefore,  the cat needs the bald eagle. \\ 
If something eats the cat and it needs the dog then the cat is cold. We know that the {\color{red}tiger} eats the cat and the cat needs the dog. Therefore,  the cat is cold. \\ 
All cold things are nice. We know that the cat is cold. Therefore,  the cat is nice. \\ 
\textbf{Answer:}  True \\
\textbf{Target Proof:} \\
If something needs the bald eagle then it eats the cat. We know that the tiger needs the bald eagle. Therefore, the tiger eats the cat. \\
If something eats the cat then the cat needs the bald eagle. We know that the tiger eats the cat. Therefore, the cat needs the bald eagle. \\ 
If something needs the bald eagle then it eats the cat. We know that the cat needs the bald eagle. Therefore, the cat eats the cat. \\
If something is red and it likes the tiger then it needs the dog. We know that the cat is red and the cat likes the tiger. Therefore, the cat needs the dog. \\
If something eats the cat and it needs the dog then the cat is cold. We know that the cat eats the cat and the cat needs the dog. Therefore, the cat is cold. \\
All cold things are nice. We know that the cat is cold. Therefore, the cat is nice. \\
\textbf{Target Answer:} True

\subsubsection*{Example 4 (value: -1.0e-05)}
\textbf{Context:}\\
If something is rough and it does not see the mouse then it does not need the bald eagle.\\
If something needs the rabbit and it likes the bald eagle then the bald eagle is blue.\\
If something sees the mouse then it needs the rabbit.\\
If something likes the rabbit and it sees the bald eagle then the bald eagle needs the rabbit.\\
If something is blue then it sees the rabbit.\\
If something likes the mouse then it is round.\\
the rabbit needs the bald eagle.\\
the rabbit likes the bald eagle.\\
the bald eagle does not see the mouse.\\
the bald eagle needs the mouse.\\
the bald eagle likes the mouse.\\
the mouse needs the bald eagle.\\
the bald eagle likes the rabbit.\\
the rabbit is red.\\
the rabbit is rough.\\
the rabbit is not green.\\
the rabbit is blue.\\
the rabbit likes the mouse.\\
the rabbit sees the mouse.\\
the bald eagle is round.\\
the rabbit sees the bald eagle.\\
the rabbit needs the mouse.\\
\textbf{Question:} Does it imply that the statement "The bald eagle sees the rabbit" is True?\\
\textbf{Proof:}\\
If something sees the mouse then it needs the rabbit. We know that the rabbit sees the mouse. Therefore,  the rabbit needs the rabbit.\\
If something sees the mouse then it needs the rabbit. We know that the rabbit sees the mouse. Therefore,  the rabbit needs the rabbit.\\
If something needs the rabbit and it likes the bald eagle then the bald eagle is blue. We know that the rabbit needs the rabbit and the rabbit likes the bald eagle. Therefore,  the bald eagle is blue.\\
If something is blue then it sees the rabbit. We know that the bald eagle is blue. Therefore,  the bald eagle sees the rabbit.\\
\textbf{Answer:}  True\\
\textbf{Target Proof:}\\
If something sees the mouse then it needs the rabbit. We know that the rabbit sees the mouse. Therefore, the rabbit needs the rabbit.\\
If something needs the rabbit and it likes the bald eagle then the bald eagle is blue. We know that the rabbit needs the rabbit and the rabbit likes the bald eagle. Therefore, the bald eagle is blue.\\
If something is blue then it sees the rabbit. We know that the bald eagle is blue. Therefore, the bald eagle sees the rabbit.\\
\textbf{Target Answer:} True\\

\subsubsection*{Example 5 (value: -1.0e-05)}

\textbf{Context:}\\
If the mouse is young and the mouse likes the cat then the mouse chases the cat.\\
If something likes the mouse and the mouse likes the tiger then the tiger likes the mouse.\\
If the tiger is cold then the tiger sees the mouse.\\
If something sees the mouse then it is young.\\
If something chases the cat and it likes the cat then the cat sees the mouse.\\
If something is young and it chases the mouse then it likes the cat.\\
If something chases the tiger and the tiger likes the cat then the cat likes the mouse.\\
If something sees the cat and it chases the cat then it likes the cat.\\
the tiger is nice.\\
the tiger sees the cat.\\
the mouse chases the tiger.\\
the tiger is red.\\
the mouse is young.\\
the mouse is round.\\
the tiger likes the mouse.\\
the mouse is cold.\\
the mouse is red.\\
the mouse likes the cat.\\
the cat chases the tiger.\\
the tiger chases the mouse.\\
\textbf{Question:}\\
Does it imply that the statement "The cat is not young" is True?\\
\textbf{Proof:}\\
If the mouse is young and the mouse likes the cat then the mouse chases the cat. We know that the mouse is young and the mouse likes the cat. Therefore,  the mouse chases the cat.\\
If the mouse is young and the mouse likes the cat then the mouse chases the cat. We know that the mouse is young and the mouse likes the cat. Therefore,  the mouse chases the cat.\\
If something chases the cat and it likes the cat then the cat sees the mouse. We know that the mouse chases the cat and the mouse likes the cat. Therefore,  the cat sees the mouse.\\
If something sees the mouse then it is young. We know that the cat sees the mouse. Therefore,  the cat is young.\\
\textbf{Answer:}  False\\
\textbf{Target Proof:}\\
If the mouse is young and the mouse likes the cat then the mouse chases the cat. We know that the mouse is young and the mouse likes the cat. Therefore, the mouse chases the cat.\\
If something chases the cat and it likes the cat then the cat sees the mouse. We know that the mouse chases the cat and the mouse likes the cat. Therefore, the cat sees the mouse.\\
If something sees the mouse then it is young. We know that the cat sees the mouse. Therefore, the cat is young.\\
\textbf{Target Answer:} False\\

\subsubsection*{Example 6 (value: -1.0e-05)}

\textbf{Context:}\\
If something eats the cow then it is green.\\
If the bald eagle eats the mouse and the mouse is nice then the mouse is rough.\\
If something needs the bald eagle then it is blue.\\
If something chases the bald eagle and the bald eagle is rough then the bald eagle eats the cow.\\
If something is green and it needs the mouse then the mouse eats the cow.\\
If something chases the bald eagle then the bald eagle is rough.\\
If something chases the dog then the dog eats the cow.\\
If something eats the dog and the dog chases the mouse then the dog needs the bald eagle.\\
If something chases the cow and it is nice then the cow chases the mouse.\\
the dog chases the cow.\\
the dog eats the mouse.\\
the cow chases the bald eagle.\\
the dog is green.\\
the cow is blue.\\
the dog chases the bald eagle.\\
the mouse needs the bald eagle.\\
the bald eagle needs the mouse.\\
the bald eagle is kind.\\
the dog chases the mouse.\\
\textbf{Question:}\\
Does it imply that the statement "The mouse is green" is True?\\
\textbf{Proof:}\\
If something is green and it needs the mouse then the mouse eats the cow. We know that {\color{red}the dog} is green and the bald eagle needs the mouse. Therefore,  the mouse eats the cow.\\
If something eats the cow then it is green. We know that the mouse eats the cow. Therefore,  the mouse is green.\\
\textbf{Answer:}  True\\
\textbf{Target Proof:}\\
If something chases the bald eagle then the bald eagle is rough. We know that the cow chases the bald eagle. Therefore, the bald eagle is rough.\\
If something chases the bald eagle and the bald eagle is rough then the bald eagle eats the cow. We know that the cow chases the bald eagle and the bald eagle is rough. Therefore, the bald eagle eats the cow.\\
If something eats the cow then it is green. We know that the bald eagle eats the cow. Therefore, the bald eagle is green.\\
If something is green and it needs the mouse then the mouse eats the cow. We know that the bald eagle is green and the bald eagle needs the mouse. Therefore, the mouse eats the cow.\\
If something eats the cow then it is green. We know that the mouse eats the cow. Therefore, the mouse is green.\\
\textbf{Target Answer:} True\\

\subsubsection*{Example 7 (value: -1.1e-05)}

\textbf{Context:}\\
If someone eats the bald eagle then the bald eagle is not kind.\\
the bear sees the cat.\\
the cat eats the bald eagle.\\
the lion sees the cat.\\
the bald eagle sees the bear.\\
the lion likes the bear.\\
the cat likes the bald eagle.\\
the bear does not eat the bald eagle.\\
\textbf{Question:}\\
Does it imply that the statement "The bald eagle is not kind" is True?\\
\textbf{Proof:}\\
If someone eats the bald eagle then the bald eagle is not kind. We know that the cat eats the bald eagle. Therefore,  the bald eagle is not kind.\\
\textbf{Answer:}  True\\
\textbf{Target Proof:}\\
If someone eats the bald eagle then the bald eagle is not kind. We know that the cat eats the bald eagle. Therefore, the bald eagle is not kind.\\
\textbf{Target Answer:} True\\

\subsubsection*{Example 8 (value: -1.1e-05)}

The question in this example asks the opposite question to that in Example 7, with the same context. Notice how the reasoning traces are the same and have the same value. The final answers are correct for each problem.

\textbf{Context:}\\
If someone eats the bald eagle then the bald eagle is not kind.\\
the bald eagle sees the bear.\\
the cat eats the bald eagle.\\
the lion sees the cat.\\
the cat likes the bald eagle.\\
the bear sees the cat.\\
the lion likes the bear.\\
the bear does not eat the bald eagle.\\
\textbf{Question:}\\
Does it imply that the statement "The bald eagle is kind" is True?\\
\textbf{Proof:}\\
If someone eats the bald eagle then the bald eagle is not kind. We know that the cat eats the bald eagle. Therefore,  the bald eagle is not kind. \\
\textbf{Answer:}  False\\
\textbf{Target Proof:}\\
If someone eats the bald eagle then the bald eagle is not kind. We know that the cat eats the bald eagle. Therefore, the bald eagle is not kind.\\
\textbf{Target Answer:} False\\

\subsubsection*{Example 9 (value: -1.2e-05)}
\textbf{Context:}\\
If someone eats the squirrel and they like the rabbit then they are green.\\
If someone sees the rabbit then they like the mouse.\\
If someone likes the squirrel then they are kind.\\
If someone likes the mouse and they like the rabbit then the mouse is red.\\
If someone likes the squirrel then the squirrel likes the rabbit.\\
If the squirrel likes the mouse and the mouse eats the squirrel then the mouse likes the squirrel.\\
If someone eats the mouse and they eat the squirrel then the mouse is green.\\
If someone is red then they eat the mouse.\\
the rabbit is red.\\
the mouse eats the rabbit.\\
the squirrel sees the mouse.\\
the mouse eats the squirrel.\\
the rabbit likes the mouse.\\
the squirrel is nice.\\
the squirrel likes the mouse.\\
the squirrel sees the rabbit.\\
the mouse likes the rabbit.\\
the rabbit sees the squirrel.\\
the squirrel eats the rabbit.\\
the rabbit eats the squirrel.\\
\textbf{Question:}\\
Does it imply that the statement "The mouse is not red" is True?\\
\textbf{Proof:}
If someone sees the rabbit then they like the mouse. We know that the squirrel sees the rabbit. Therefore,  the squirrel likes the mouse. \\
If someone likes the mouse and they like the rabbit then the mouse is red. We know that the squirrel likes the mouse and the {\color{red}mouse} likes the rabbit. Therefore,  the mouse is red. \\
\textbf{Answer:}  False\\
\textbf{Target Proof:}\\
If the squirrel likes the mouse and the mouse eats the squirrel then the mouse likes the squirrel. We know that the squirrel likes the mouse and the mouse eats the squirrel. Therefore, the mouse likes the squirrel. \\
If someone likes the squirrel then the squirrel likes the rabbit. We know that the mouse likes the squirrel. Therefore, the squirrel likes the rabbit.\\
If someone likes the mouse and they like the rabbit then the mouse is red. We know that the squirrel likes the mouse and the squirrel likes the rabbit. Therefore, the mouse is red.\\
Target Answer: False\\

\subsubsection*{Example 10 (value: -1.2e-05)}
\textbf{Context:}
If someone likes the mouse and the mouse sees the cow then they like the dog. \\
If someone is green and they chase the lion then the lion chases the mouse.\\
If the mouse does not like the lion then the mouse does not chase the lion.\\
If the dog is young then the dog is round.\\
If someone likes the lion then the lion likes the cow.\\
If someone is green then they like the cow.\\
If someone likes the cow and they are rough then the cow does not like the lion.\\
If someone likes the cow then they chase the cow.\\
the lion likes the cow.\\
the cow sees the lion.\\
the dog sees the mouse.\\
the mouse chases the cow.\\
the dog is green.\\
the mouse does not chase the dog.\\
the mouse is rough.\\
\textbf{Question:} \\
Does it imply that the statement "The dog does not like the cow" is True? \\
\textbf{Proof:}\\
If someone is green then they like the cow. We know that the dog is green. Therefore,  the dog likes the cow. \\
\textbf{Answer:}  False \\
\textbf{Target Proof:}\\
If someone is green then they like the cow. We know that the dog is green. Therefore, the dog likes the cow.\\
\textbf{Target Answer:} False\\

Below are examples of the worst five reasoning traces from \textbf{SI + Halter + Search} according to the value function. Errors are highlighted in red.

\subsubsection*{Example 1 (value: -10.9)}

\textbf{Context:}\\
If something is kind then it likes the cow.\\
If something likes the cow then the cow is kind.\\
the cow is big.\\
the mouse eats the bear.\\
the tiger is kind.\\
the bear visits the tiger.\\
\textbf{Question:}\\
Does it imply that the statement "The cow likes the cow" is True? \\
\textbf{Proof:}\\
If something is kind then it likes the cow. We know that the tiger is kind. Therefore,  the tiger likes the cow.\\
If something likes the cow then the cow is kind. We know that the tiger likes the cow. Therefore,  the cow is kind.\\
If something is kind then it likes the cow. We know that the cow is kind. Therefore,  the cow likes the cow.\\
\textbf{Answer: } True\\
\textbf{Target Proof:}
If something is kind then it likes the cow. We know that the tiger is kind. Therefore, the tiger likes the cow.\\
If something likes the cow then the cow is kind. We know that the tiger likes the cow. Therefore, the cow is kind.\\
If something is kind then it likes the cow. We know that the cow is kind. Therefore, the cow likes the cow.\\
\textbf{Target Answer:} True\\

\subsubsection*{Example 2 (value: -10.2)}
\textbf{Context:}\\
If something needs the squirrel then the squirrel is cold.\\
If something needs the lion then the lion likes the cat.\\
If the lion needs the squirrel and the squirrel needs the lion then the lion eats the squirrel.\\
If something likes the squirrel and the squirrel likes the cow then the cow is rough.\\
If something is cold then it needs the squirrel.\\
If something needs the squirrel and it eats the cat then the cat likes the squirrel.\\
the cow is cold.\\
the cat eats the cow.\\
the lion eats the cow.\\
the squirrel eats the lion.\\
\textbf{Question:}\\
Does it imply that the statement "The squirrel does not need the squirrel" is True?\\
\textbf{Proof:}\\
If something is cold then it needs the squirrel. We know that the cow is cold. Therefore,  the cow needs the squirrel.\\
If something needs the squirrel then the squirrel is cold. We know that the cow needs the squirrel. Therefore,  the squirrel is cold.\\
If something is cold then it needs the squirrel. We know that the squirrel is cold. Therefore,  the squirrel needs the squirrel.\\
\textbf{Answer:}  False\\
\textbf{Target Proof:}\\
If something is cold then it needs the squirrel. We know that the cow is cold. Therefore, the cow needs the squirrel.\\
If something needs the squirrel then the squirrel is cold. We know that the cow needs the squirrel. Therefore, the squirrel is cold.\\
If something is cold then it needs the squirrel. We know that the squirrel is cold. Therefore, the squirrel needs the squirrel.\\
\textbf{Target Answer:} False\\

\subsubsection*{Example 3 (value: -10.2)}

\textbf{Context:}\\
If someone is kind then they are rough.\\
All cold, quiet people are smart.\\
If someone is cold then they are smart.\\
If someone is blue then they are quiet.\\
If someone is red and kind then they are smart.\\
If someone is quiet then they are blue.\\
All blue, smart people are red.\\
Blue, rough people are red.\\
If Gary is kind and Gary is rough then Gary is quiet.\\
Fiona is smart.\\
Charlie is kind.\\
Fiona is blue.\\
Bob is smart.\\
Fiona is rough.\\
Gary is kind.\\
Charlie is not smart.\\
\textbf{Question:}\\
Does it imply that the statement "Gary is smart" is True? \\
\textbf{Proof:}\\
If someone is kind then they are rough. We know that Gary is kind. Therefore,  Gary is rough. \\
If someone is red and kind then they are smart. We know that Gary is kind and {\color{red}Gary is kind}. Therefore,  Gary is smart.\\
\textbf{Answer:} True\\
\textbf{Target Proof:} \\
If someone is kind then they are rough. We know that Gary is kind. Therefore, Gary is rough.\\
If Gary is kind and Gary is rough then Gary is quiet. We know that Gary is kind and Gary is rough. Therefore, Gary is quiet.\\
If someone is quiet then they are blue. We know that Gary is quiet. Therefore, Gary is blue.\\
If someone is kind then they are rough. We know that Gary is kind. Therefore, Gary is rough.\\
Blue, rough people are red. We know that Gary is blue and Gary is rough. Therefore, Gary is red.\\
If someone is red and kind then they are smart. We know that Gary is red and Gary is kind. Therefore, Gary is smart.\\
\textbf{Target Answer:} True \\

\subsubsection*{Example 4 (value: -10.1)}
\textbf{Context:}\\
If something likes the cow then the cow is kind.\\
If something is kind then it likes the cow.\\
the cow is big.\\
the tiger is kind.\\
the mouse eats the bear.\\
the bear visits the tiger.\\
\textbf{Question:}\\
Does it imply that the statement "The cow does not like the cow" is True? \\
\textbf{Proof:}\\
If something is kind then it likes the cow. We know that the tiger is kind. Therefore,  the tiger likes the cow.\\
If something likes the cow then the cow is kind. We know that the tiger likes the cow. Therefore,  the cow is kind.\\
If something is kind then it likes the cow. We know that the cow is kind. Therefore,  the cow likes the cow.\\
\textbf{Answer:}  False\\
\textbf{Target Proof:}\\
If something is kind then it likes the cow. We know that the tiger is kind. Therefore, the tiger likes the cow.\\
If something likes the cow then the cow is kind. We know that the tiger likes the cow. Therefore, the cow is kind.\\
If something is kind then it likes the cow. We know that the cow is kind. Therefore, the cow likes the cow.\\
\textbf{Target Answer:} False\\

\subsubsection*{Example 5 (value: -10.0)}

\textbf{Context:}\\
If something is kind and not young then it eats the mouse.\\
If something eats the mouse and the mouse eats the tiger then the mouse does not see the rabbit.\\
If something needs the rabbit and the rabbit eats the tiger then the tiger sees the rabbit.\\
If something is blue then it eats the mouse.\\
If something eats the mouse then the mouse is blue.\\
If something sees the tiger then it eats the rabbit.\\
If the mouse is young and the mouse eats the bald eagle then the mouse needs the rabbit.\\
If something sees the rabbit then the rabbit is not blue.\\
the tiger is green.\\
the bald eagle sees the tiger.\\
the tiger is blue.\\
the bald eagle is not blue.\\
the rabbit is round.\\
the rabbit is kind.\\
the mouse is not kind.\\
the bald eagle eats the rabbit.\\
the tiger sees the rabbit.\\
the bald eagle sees the mouse.\\
\textbf{Question:}\\
Does it imply that the statement "The mouse does not eat the mouse" is True?\\
\textbf{Proof:}\\
If something is blue then it eats the mouse. We know that the tiger is blue. Therefore,  the tiger eats the mouse.\\
If something eats the mouse then the mouse is blue. We know that the tiger eats the mouse. Therefore,  the mouse is blue.\\
If something is blue then it eats the mouse. We know that the mouse is blue. Therefore,  the mouse eats the mouse.\\
\textbf{Answer:}  False\\
\textbf{Target Proof:}\\
If something is blue then it eats the mouse. We know that the tiger is blue. Therefore, the tiger eats the mouse.\\
If something eats the mouse then the mouse is blue. We know that the tiger eats the mouse. Therefore, the mouse is blue.\\
If something is blue then it eats the mouse. We know that the mouse is blue. Therefore, the mouse eats the mouse.\\
\textbf{Target Answer:} False\\

\subsection{EntailmentBankQA Task 1 - depth 1: SI + Halter + Search}

Below are the top five examples of reasoning traces on EntailmentBankQA problems that require a single step of reasoning, ranked by their value according to our learned value function and are otherwise not cherry picked.

\subsubsection*{Example 1 (value: -0.026)}
\textbf{Context:}\\
a fly is a kind of insect\\
an insect has six legs\\
\textbf{Question:}\\
An animal has six legs. What is it most likely to be? a spider OR a fly OR a lizard OR a centipede.\\
\textbf{Proof:}\\
a fly is a kind of insect. We know that an insect has six legs. Therefore,  a fly has six legs.\\
\textbf{Answer: } a fly\\
\textbf{Target Proof:}\\
a fly is a kind of insect. We know that an insect has six legs. Therefore, a fly has six legs.\\
\textbf{Target Answer:} a fly\\

\subsubsection*{Example 2 (value: -0.033)}
\textbf{Context:}\\
solid is a kind of state of matter\\
an ice cube is a kind of solid\\
\textbf{Question:}\\
Which word best describes the physical state of an ice cube? gas OR solid OR liquid OR plasma.\\
\textbf{Proof:}\\
solid is a kind of state of matter. We know that an ice cube is a kind of solid. Therefore,  an ice cube is in solid state.\\
\textbf{Answer:}  solid\\
\textbf{Target Proof}:\\
solid is a kind of state of matter. We know that an ice cube is a kind of solid. Therefore, an ice cube is solid in its physical state.\\
\textbf{Target Answer: } solid\\

\subsubsection*{Example 3 (value: -0.045)}
\textbf{Context:}\\
the surface of the moon contains mountains\\
the surface of the earth contains mountains\\
\textbf{Question:}\\
Which features can be found on the surface of both Earth and the Moon? plants OR oceans OR animals OR mountains.\\
\textbf{Proof:}\\
the surface of the moon contains mountains. We know that the surface of the earth contains mountains. Therefore,  both the surface of the moon and the surface of the earth contain mountains.\\
\textbf{Answer: } mountains\\
\textbf{Target Proof:}\\
the surface of the moon contains mountains. We know that the surface of the earth contains mountains. Therefore, both the surface of the moon and the surface of the earth contain mountains.\\
\textbf{Target Answer:} mountains\\

\subsubsection*{Example 4 (value: -0.049)}
\textbf{Context:}\\
steel is a kind of solid metal \\
sound travels fastest through solid \\
\textbf{Question:}\\
Through which material would sound travel the fastest? cork OR water OR air OR steel.\\
\textbf{Proof:}\\
steel is a kind of solid metal. We know that sound travels fastest through solid. Therefore,  sound will travel fastest in steel.\\
\textbf{Answer:}  steel\\
\textbf{Target Proof:}\\
steel is a kind of solid metal. We know that sound travels fastest through solid. Therefore, sound travels fastest in steel.\\
\textbf{Target Answer:} steel\\

\subsubsection*{Example 5 (value: -0.052)}
\textbf{Context:}\\
drinking straws are usually made of plastic\\
plastic is flexible\\
\textbf{Question:}\\
Which of these is MOST flexible? Broom handle OR Wooden ruler OR Drinking straw OR Sewing needle.\\
\textbf{Proof:}\\
drinking straws are usually made of plastic. We know that plastic is flexible. Therefore,  drinking straws are flexible.\\
\textbf{Answer:}  Drinking straw\\
\textbf{Target Proof:}\\
drinking straws are usually made of plastic. We know that plastic is flexible. Therefore, a drinking straw is flexible.\\
\textbf{Target Answer:} Drinking straw\\

\subsection{EntailmentBankQA Task 1 - depth 2: SI + Halter + Search}

Below are the top five examples of reasoning traces on EntailmentBankQA problems that require two steps of reasoning, ranked by their value according to our learned value function and are otherwise not cherry picked.

\subsubsection*{Example 1 (value: -0.022)}
\textbf{Context:}\\
a runway is a kind of pathway for airplanes\\
airports have runways for airplanes\\
as the number of pathways increases , the traffic congestion in that area usually decreases\\
\textbf{Question:}\\
Which of the following would be most effective in reducing air traffic congestion at a busy airport? providing performance feedback to pilots OR providing flight information to passengers OR increasing the number of aircraft at the airport OR increasing the number of runways at the airport.\\
\textbf{Proof:}\\
a runway is a kind of pathway for airplanes. We know that airports have runways for airplanes. Therefore,  an airport runway is a kind of pathway for airplanes.\\
an airport runway is a kind of pathway for airplanes. We know that as the number of pathways increases , the traffic congestion in that area usually decreases. Therefore,  as the number of runways at a airport increases, the traffic congestion in that area usually decreases.\\
\textbf{Answer:}  increasing the number of runways at the airport\\
\textbf{Target Proof:}\\
a runway is a kind of pathway for airplanes. We know that as the number of pathways increases , the traffic congestion in that area usually decreases. Therefore, as the number of runways increases, the traffic congestion in that area usually decreases.\\
as the number of runways increases, the traffic congestion in that area usually decreases. We know that airports have runways for airplanes. Therefore, as the number of runways increases, the traffic congestion at the airport usually decreases.\\
\textbf{Target Answer:} increasing the number of runways at the airport\\

\subsubsection*{Example 2 (value: -0.030)}
\textbf{Context:}\\
tidal energy is a renewable resource\\
wave energy is a renewable resource\\
tidal energy means energy from ocean tides\\
\textbf{Question:}\\
Wave energy from the ocean can be harnessed to power generators to make electricity. Energy from ocean tides can also be used to make electricity. How would you categorize these two sources of energy? Both are renewable. OR Both are nonrenewable. OR Wave energy is nonrenewable. Tidal energy is renewable. OR Wave energy is renewable. Tidal energy is nonrenewable..\\
\textbf{Proof:}\\
tidal energy is a renewable resource. We know that wave energy is a renewable resource. Therefore,  wave energy and tidal energy are both renewable resources.\\
Answer:  Wave energy is renewable \\
\textbf{Target Proof}:\\
tidal energy is a renewable resource. We know that tidal energy means energy from ocean tides. Therefore, energy from ocean tides is a renewable resource.\\
energy from ocean tides is a renewable resource. We know that wave energy is a renewable resource. Therefore, wave energy and energy from ocean tides are both renewable resources.\\
Target Answer: Both are renewable.\\

\subsubsection*{Example 3 (value: -0.031}
\textbf{Context:}
light comes to the focal point after passing through a convex lens\\
a convex lens causes light to refract and concentrate to magnify distant objects\\
refraction is when light bends\\
\textbf{Question:}\\
Which object is best used to demonstrate the bending of light as it passes through a substance? convex lens OR flat mirror OR opaque rod OR concave mirror.\\
\textbf{Proof:}\\
light comes to the focal point after passing through a convex lens. We know that a convex lens causes light to refract and concentrate to magnify distant objects. Therefore,  the lens causes light to refract and concentrate on the focal point.\\
the lens causes light to refract and concentrate on the focal point. We know that refraction is when light bends. Therefore,  the lens causes the light to bend towards the focal point.\\
\textbf{Answer: } convex lens\\
\textbf{Target Proof:}\\
light comes to the focal point after passing through a convex lens. We know that a convex lens causes light to refract and concentrate to magnify distant objects. Therefore, a convex lens will cause the light to refract when the light passes through the convex lens.\\
a convex lens will cause the light to refract when the light passes through the convex lens. We know that refraction is when light bends. Therefore, a convex lens will cause the light to bend when the light passes through the lens.\\
\textbf{Target Answer:} convex lens\\

\subsubsection*{Example 4 (value: -0.036)}
\textbf{Context:}\\
friction occurs when two object 's surfaces move against each other\\
a hand is a kind of object\\
friction causes the temperature of an object to increase\\
a student is rubbing his hands together\\
\textbf{Question:}\\
A student stands outside on a cold winter day. His hands become cold and he rubs them together to make them warmer. Which statement explains why rubbing his hands together makes them warmer? This action produces thermal energy through friction. OR This action conducts thermal energy away from the body. OR This action captures thermal energy from the environment. OR This action reduces the amount of thermal energy transferred to the air..\\
\textbf{Proof:}\\
friction occurs when two object 's surfaces move against each other. We know that a hand is a kind of object and a student is rubbing his hands together. Therefore,  friction will occur between the student's hands.\\
friction will occur between the student's hands. We know that friction causes the temperature of an object to increase. Therefore,  friction will cause the temperature of the student's hands to increase.\\
\textbf{Answer:}  This action produces thermal energy through friction\\
\textbf{Target Proof:}\\
friction occurs when two object 's surfaces move against each other. We know that a hand is a kind of object and a student is rubbing his hands together. Therefore, friction occurs when the student is rubbing his hands together.\\
friction occurs when the student is rubbing his hands together. We know that friction causes the temperature of an object to increase. Therefore, friction causes the temperature of student's hands to increase.\\
\textbf{Target Answer:} This action produces thermal energy through friction.\\

\subsubsection*{Example 5 (value: -0.0378)}
\textbf{Context:}
receiving the recessive gene for a trait from each parent causes the offspring to express that recessive trait\\
light eyes are a recessive trait for human\\
jessica is a human with light eyes\\
\textbf{Question:}\\
If Jessica has light eyes (bb) and both of her parents have dark eyes (Bb), which statement is true? Jessica inherited both genes from her father. OR Jessica inherited both genes from her mother. OR Jessica inherited one recessive form of the gene from each parents from each parent. OR Jessica inherited one dominant from of the gene from each parent..\\
\textbf{Proof:}
receiving the recessive gene for a trait from each parent causes the offspring to express that recessive trait. We know that light eyes are a recessive trait for human and jessica is a human with light eyes. Therefore,  jessica received the recessive gene for light eyes from each parent.\\
\textbf{Answer:}  Jessica inherited one recessive form of the gene from each parent\\
\textbf{Target Proof:}\\
light eyes are a recessive trait for human. We know that jessica is a human with light eyes. Therefore, jessica has a recessive trait of light eyes.\\
jessica has a recessive trait of light eyes. We know that receiving the recessive gene for a trait from each parent causes the offspring to express that recessive trait. Therefore, jessica received the recessive gene from each of her parent.\\
\textbf{Target Answer:} Jessica inherited one recessive form of the gene from each parents from each parent.\\

\subsection{EntailmentBankQA Task 1 - depth 3: SI + Halter + Search}

Below are the top five examples of reasoning traces on EntailmentBankQA problems that require three steps of reasoning, ranked by their value according to our learned value function and are otherwise not cherry picked.

\subsubsection*{Example 1 (value: -0.0080)}
\textbf{Context:}\\
if something receives sunlight, then it will increase in temperature\\
if something increases in temperature, then that something will be warm\\
the air around the equator receives direct sunlight\\
the air mass formed close to the ocean is humid in moisture\\
\textbf{Question:}\\
Which type of air mass forms over the ocean near the equator? moist and warm OR moist and cool OR dry and warm OR dry and cool.\\
\textbf{Proof:}\\
the air around the equator receives direct sunlight. We know that the air mass formed close to the ocean is humid in moisture. Therefore,  the air mass formed close to the ocean would receive direct sunlight.\\
the air mass formed close to the ocean would receive direct sunlight. We know that if something receives sunlight, then it will increase in temperature. Therefore,  the air mass formed close to the ocean would increase in temperature.\\
the air mass formed close to the ocean would increase in temperature. We know that if something increases in temperature, then that something will be warm. Therefore,  the air mass formed close to the ocean would be warm.\\
\textbf{Answer:}  moist and warm\\
\textbf{Target Proof:}\\
if something receives sunlight, then it will increase in temperature. We know that the air around the equator receives direct sunlight. Therefore, the air around the equator will increase in temperature.\\
the air around the equator will increase in temperature. We know that if something increases in temperature, then that something will be warm. Therefore, the air is warm around the equator.\\
the air is warm around the equator. We know that the air mass formed close to the ocean is humid in moisture. Therefore, air masses formed over the ocean near the equator are moist / warm.\\
\textbf{Target Answer:} moist and warm\\

\subsubsection*{Example 2 (value: -0.014)}
\textbf{Context:}
sliding is a kind of movement\\
friction occurs when two objects move against each other\\
a rock is a kind of object\\
a frozen pond is a kind of object\\
a rock is slided across the frozen pond\\
friction causes the speed of an object to decrease\\
\textbf{Question:}\\
Kelly slides a flat rock across the smooth ice of a frozen pond. The rock slows down after several seconds. What causes the rock to slow down? The thickness of the ice OR The temperature of the air above the ice OR The force of friction between the ice and the rock OR The gravitational force between the ice and the rock.\\
\textbf{Proof:}\\
sliding is a kind of movement. We know that a rock is slided across the frozen pond and a rock is a kind of object. Therefore,  the rock is slid across the frozen pond.\\
the rock is slid across the frozen pond. We know that friction occurs when two objects move against each other and a rock is a kind of object and a frozen pond is a kind of object. Therefore,  friction occurs when the rock is slid across the frozen pond.\\
friction occurs when the rock is slid across the frozen pond. We know that friction causes the speed of an object to decrease and a rock is a kind of object. Therefore,  friction will cause the speed of the rock to decrease when it is slid across the frozen pond.\\
\textbf{Answer: } The force of friction between the ice and the rock\\
\textbf{Target Proof:}\\
sliding is a kind of movement. We know that a rock is slided across the frozen pond. Therefore, a rock is moving on the frozen pond.\\
a rock is moving on the frozen pond. We know that friction occurs when two objects move against each other and a rock is a kind of object and a frozen pond is a kind of object. Therefore, friction will occur between the rock and the  frozen pond.\\
friction will occur between the rock and the  frozen pond. We know that a rock is a kind of object and friction causes the speed of an object to decrease. Therefore, friction will cause the speed of the rock to decrease.\\
\textbf{Target Answer:} The force of friction between the ice and the rock\\

\subsubsection*{Example 3 (value: -0.027)}
\textbf{Context:}\\
the parent cats have white fur\\
white fur is white in color\\
inheriting is when an inherited characteristic is passed from parent to offspring\\
a cat is a kind of animal\\
the color of / coloration of fur is an inherited characteristic\\
animals produce offspring\\
\textbf{Question:}\\
Which trait would a cat most likely inherit from its parents? having a scar OR chasing a toy OR having white fur OR knowing its name.\\
\textbf{Proof:}\\
inheriting is when an inherited characteristic is passed from parent to offspring. We know that the color of / coloration of fur is an inherited characteristic. Therefore,  offspring can inherit the color of fur from parents.\\
offspring can inherit the color of fur from parents. We know that the parent cats have white fur and white fur is white in color. Therefore,  the offspring of the cats will inherit white fur from their parents.\\
\textbf{Answer:}  having white fur\\
\textbf{Target Proof:}\\
inheriting is when an inherited characteristic is passed from parent to offspring. We know that the color of / coloration of fur is an inherited characteristic. Therefore, the offspring will inherit the color of the fur of its parent.\\
the offspring will inherit the color of the fur of its parent. We know that a cat is a kind of animal and animals produce offspring. Therefore, a cat will inherit the color of the fur of its parents.\\
a cat will inherit the color of the fur of its parents. We know that the parent cats have white fur and white fur is white in color. Therefore, the cat will inherit the white colored fur from its parents.\\
\textbf{Target Answer:} having white fur\\

\subsubsection*{Example 4 (value: -0.027)}
\textbf{Context:}\\
carbon is a kind of element\\
a compound is made of two or more elements chemically combined\\
calcium carbonate is made of calcium, carbon, and oxygen\\
calcium is a kind of element\\
oxygen is a kind of element\\
\textbf{Question:}\\
Calcium carbonate (CaCO$_{3}$) is considered a compound because it is made up of two or more molecules. OR elements. OR gases. OR solids..\\
\textbf{Proof:}\\
carbon is a kind of element. We know that calcium is a kind of element and oxygen is a kind of element. Therefore,  calcium, oxygen, and carbon are all elements.\\
calcium, oxygen, and carbon are all elements. We know that calcium carbonate is made of calcium, carbon, and oxygen. Therefore,  calcium carbonate is made of elements.\\
\textbf{Answer:}  elements\\
\textbf{Target Proof:}\\
carbon is a kind of element. We know that calcium is a kind of element and oxygen is a kind of element. Therefore, calcium, carbon, and oxygen are three elements.\\
calcium, carbon, and oxygen are three elements. We know that calcium carbonate is made of calcium, carbon, and oxygen. Therefore, calcium carbonate is made of three elements.\\
calcium carbonate is made of three elements. We know that a compound is made of two or more elements chemically combined. Therefore, calcium carbonate is a kind of compound.\\
\textbf{Target Answer:} elements.\\

\subsubsection*{Example 5 (value: -0.031)}
\textbf{Context:}\\
trees takes in oxygen through roots close to the surface of the ground\\
earthworms create tunnels in soil\\
the looseness of soil increases the amount of oxygen in that soil\\
tunnels in soil loosen that soil\\
\textbf{Question:}\\
Trees need oxygen. Roots close to the surface of the ground take in the oxygen the tree needs. Which organisms help trees get oxygen? woodpeckers making holes in the tree OR earthworms making holes in the ground near the tree OR mushrooms growing at the base of the tree OR squirrels eating walnuts on the ground near the tree.\\
\textbf{Proof:}\\
tunnels in soil loosen that soil. We know that the looseness of soil increases the amount of oxygen in that soil. Therefore,  tunnels in soil increase the amount of oxygen in the soil.\\
tunnels in soil increase the amount of oxygen in the soil. We know that trees takes in oxygen through roots close to the surface of the ground. Therefore,  tunnels in soil increase the amount of oxygen that can be taken in through the roots of trees.\\
tunnels in soil increase the amount of oxygen that can be taken in through the roots of trees. We know that earthworms create tunnels in soil. Therefore,  earthworms help trees get more oxygen through their roots by creating tunnels in the soil.\\
\textbf{Answer:}  earthworms making holes in the ground near the tree\\
\textbf{Target Proof:}\\
earthworms create tunnels in soil. We know that tunnels in soil loosen that soil. Therefore, earthworm can help loosen the soil by creating tunnels.\\
earthworm can help loosen the soil by creating tunnels. We know that the looseness of soil increases the amount of oxygen in that soil. Therefore, earthworm creating tunnes can increase the amount of oxygen in the soil.\\
earthworm creating tunnes can increase the amount of oxygen in the soil. We know that trees takes in oxygen through roots close to the surface of the ground. Therefore, more oxygen is available for tree roots.\\
\textbf{Target Answer:} earthworms making holes in the ground near the tree\\

\subsection{EntailmentBankQA Task 1 - depth 5: SI + Halter + Search}

Below are the top five examples of reasoning traces on EntailmentBankQA problems that require a five steps of reasoning, ranked by their value according to our learned value function and are otherwise not cherry picked.

\subsubsection*{Example 1 (value: -0.033)}
\textbf{Context:}\\
woodpeckers live in forests\\
animals adapt to their environment\\
a squirrel is a kind of animal\\
a woodpecker is a kind of bird\\
squirrels live in forests\\
a deer lives in a forest\\
a deer is a kind of animal\\
a forest is a kind of environment\\
\textbf{Question:}\\
In which group of animals below are all of the members well adapted to living in a forest ecosystem? field mouse, rabbit, pheasant OR gull, tree frog, grasshopper OR coyote, antelope, bison OR squirrel, deer, woodpecker.\\
\textbf{Proof:}\\
a squirrel is a kind of animal. We know that a deer is a kind of animal. Therefore,  a deer and a squirrel are kinds of animals.\\
a deer and a squirrel are kinds of animals. We know that animals adapt to their environment. Therefore,  a deer and a squirrel adapt to their environment.\\
\textbf{Answer:}  squirrel, deer, woodpecker \\
\textbf{Target Proof:}\\
a squirrel is a kind of animal. We know that squirrels live in forests. Therefore, squirrels are animals live in forests.\\
a deer lives in a forest. We know that a deer is a kind of animal. Therefore, deer is an animal lives in a forest.\\
woodpeckers live in forests. We know that a woodpecker is a kind of bird. Therefore, woodpeckers are animals live in forests.\\
squirrels are animals live in forests. We know that deer is an animal lives in a forest and woodpeckers are animals live in forests.\\ Therefore, squirrels and deer and woodpeckers are animals that live in forests.\\
squirrels and deer and woodpeckers are animals that live in forests. We know that animals adapt to their environment and a forest is a kind of environment. Therefore, squirrels and deer and woodpeckers are adapted to living in forests.\\
\textbf{Target Answer:} squirrel, deer, woodpecker\\

\subsubsection*{Example 2 (value: -0.039)}
\textbf{Context:}\\
toxic chemicals are dumped into a pond\\
poisonous things can cause harm to organisms\\
toxic chemicals are poisonous\\
harming an organism can cause that organism to die\\
a fish is a kind of organism\\
a pond is a kind of body of water\\
a fish lives in water\\
\textbf{Question:}\\
Dumping toxic chemicals into a pond would most likely result in an increase in oxygen levels in the pond. OR plants near the pond growing more quickly. OR the toxic chemicals having no effect on the pond. OR fish in the pond being harmed or dying off..\\
\textbf{Proof:}\\
a fish is a kind of organism. We know that harming an organism can cause that organism to die. Therefore,  harming a fish can cause that fish to die.\\
harming a fish can cause that fish to die. We know that poisonous things can cause harm to organisms. Therefore,  poisonous things can cause fish to die.\\
harming a fish can cause that fish to die. We know that poisonous things can cause fish to die. Therefore,  poisonous things can harm fish.\\
poisonous things can cause fish to die. We know that toxic chemicals are poisonous. Therefore,  toxic chemicals can cause fish to die.\\
\textbf{Answer:}  fish in the pond being harmed or dying off\\
\textbf{Target Proof:}\\
toxic chemicals are dumped into a pond. We know that toxic chemicals are poisonous. Therefore, the pond is poisonous because of toxic chemicals.\\
poisonous things can cause harm to organisms. We know that harming an organism can cause that organism to die. Therefore, poisonous things can cause organisms to die.\\
a pond is a kind of body of water. We know that a fish lives in water. Therefore, fish live in ponds.\\
fish live in ponds. We know that a fish is a kind of organism. Therefore, a fish is a kind of organism that lives in ponds.\\
the pond is poisonous because of toxic chemicals. We know that poisonous things can cause organisms to die and a fish is a kind of organism that lives in ponds. Therefore, the pond with toxic chemicals can cause fish in the pond to die.\\
\textbf{Target Answer: }fish in the pond being harmed or dying off.\\

\subsubsection*{Example 3 (value: -0.049)}

In this example, the model makes a mistake on the first step of reasoning but is  able to recover and does not use the incorrect inference in later steps.

\textbf{Context:}\\
pulling a plow is a kind of chore\\
a horse is a kind of animal\\
an instinctive behavior is a kind of inherited characteristic\\
inherited characteristics are the opposite of learned characteristics\\
doing chores is a learned characteristic\\
a learned characteristic is a kind of characteristic of an animal\\
\textbf{Question:}\\
Which of these is not an instinctive behavior? a bird building a nest OR a turtle burying its eggs OR a bear hibernating in winter OR a horse pulling a plow.\\
\textbf{Proof:}\\
{\color{red} doing chores is a learned characteristic. We know that inherited characteristics are the opposite of learned characteristics. Therefore, \textbf{doing chores is an inherited characteristic}.}\\
a horse is a kind of animal. We know that a learned characteristic is a kind of characteristic of an animal. Therefore,  a learned characteristic of a horse is a kind of learned characteristic of an animal.\\
pulling a plow is a kind of chore. We know that doing chores is a learned characteristic. Therefore,  pulling a plow is a kind of learned characteristic.\\
a learned characteristic of a horse is a kind of learned characteristic of an animal. We know that inherited characteristics are the opposite of learned characteristics. Therefore,  a horse cannot inherit a learned characteristic.\\
pulling a plow is a kind of learned characteristic. We know that a horse cannot inherit a learned characteristic. Therefore,  a horse cannot inherit the characteristic of pulling a plow.\\
\textbf{Answer:}  a horse pulling a plow\\
\textbf{Target Proof:}\\
pulling a plow is a kind of chore. We know that doing chores is a learned characteristic. Therefore, pulling a plow is a learned characteristic.\\
pulling a plow is a learned characteristic. We know that a learned characteristic is a kind of characteristic of an animal. Therefore, an animal pulling a plow is a kind of learned characteristic.\\
an animal pulling a plow is a kind of learned characteristic. We know that a horse is a kind of animal. Therefore, a horse pulling a plow is a learned characteristic.\\
an instinctive behavior is a kind of inherited characteristic. We know that inherited characteristics are the opposite of learned characteristics. Therefore, instinctive behaviors are opposite to learned characteristics.\\
a horse pulling a plow is a learned characteristic. We know that instinctive behaviors are opposite to learned characteristics. Therefore, a horse pulling a plow is not an instinctive behavior.\\
\textbf{Target Answer:} a horse pulling a plow\\

\subsubsection*{Example 4 (value: -0.052)}
\textbf{Context:}\\
disease-causing microbes have a negative impact on the body\\
decreasing something negative has a positive impact on a thing\\
destroying something decreases the amount of that something\\
a source of something produces that something\\
protecting a living thing has a positive impact on that living thing 's survival / health\\
specialized cells are a source of chemicals that destroy\\ disease-causing microbes\\
\textbf{Question:}\\
Specialized cells protect the human body from disease-causing microbes by producing chemicals that destroy the microbes OR dissolving wastes from the microbes OR creating recessive genes in the microbes OR supplying oxygen and nutrients to the microbes.\\
\textbf{Proof:}\\
disease-causing microbes have a negative impact on the body. We know that decreasing something negative has a positive impact on a thing. Therefore,  decreasing disease-causing microbes has a positive impact on the body.\\
decreasing disease-causing microbes has a positive impact on the body. We know that specialized cells are a source of chemicals that destroy disease-causing microbes. Therefore,  as the number of specialized cells in the body increases, the number of disease-causing microbes being destroyed will increase.\\
\textbf{Answer:}  {\color{red}creating recessive genes in the microbes}\\
\textbf{Target Proof:}\\
a source of something produces that something. We know that specialized cells are a source of chemicals that destroy disease-causing microbes. Therefore, specialized cells produce chemicals that destroy disease-causing microbes.\\
decreasing something negative has a positive impact on a thing. We know that destroying something decreases the amount of that something. Therefore, destroying something negative has a positive impact on a thing.\\
destroying something negative has a positive impact on a thing. We know that disease-causing microbes have a negative impact on the body. Therefore, destroying disease-causing microbes has a positive impact on the body.\\
destroying disease-causing microbes has a positive impact on the body. We know that protecting a living thing has a positive impact on that living thing 's survival / health. Therefore, destroying disease-causing microbes is a kind of method for protecting the body.\\
specialized cells produce chemicals that destroy disease-causing microbes. We know that destroying disease-causing microbes is a kind of method for protecting the body. Therefore, specialized cells protect the body be producing chemicals that destroy disease-causing microbes.\\
\textbf{Target Answer:} producing chemicals that destroy the microbes\\

\subsubsection*{Example 5 (value: -0.057)}
\textbf{Context:}\\
building materials are used to build buildings\\
wood boards can be used to build houses\\
a house is a kind of building\\
wood is a renewable resource\\
wood boards are made of wood\\
a renewable resource is a kind of natural resource\\
\textbf{Question:}\\
The building of houses in Maryland uses many different materials. Which building material is made of a renewable natural resource? copper wire OR steel beams OR wood boards OR plastic siding.\\
\textbf{Proof:}\\
wood boards can be used to build houses. We know that wood is a renewable resource. Therefore,  wood boards are a renewable resource that can be used to build houses.\\
\textbf{Answer:}  wood boards\\
\textbf{Target Proof:}\\
building materials are used to build buildings. We know that a house is a kind of building. Therefore, building materials can be used to build houses.\\
building materials can be used to build houses. We know that wood boards can be used to build houses. Therefore, wood boards are a kind of building material.\\
wood is a renewable resource. We know that a renewable resource is a kind of natural resource. Therefore, wood is a kind of renewable natural resource.\\
wood is a kind of renewable natural resource. We know that wood boards are made of wood. Therefore, wood boards are made of a renewable natural resource.\\
wood boards are a kind of building material. We know that wood boards are made of a renewable natural resource. Therefore, wood boards are a kind of building material that is made of a renewable natural resource.\\
\textbf{Target Answer:} wood boards\\

\subsection{EntailmentBankQA Task 2 - depth 1: SI + Halter + Search}

Below are the top three examples of reasoning traces on EntailmentBankQA problems that require a single step of reasoning, ranked by their value according to our learned value function and are otherwise not cherry picked.

\subsubsection*{Example 1 (value: -0.018)}
\textbf{Context:}
oil is a kind of energy resource\\
gas is a kind of substance\\
gasoline is a source of chemical energy\\
chemical energy is a kind of potential energy\\
an engine is a part of an automobile\\
a fuel is used to make energy\\
ethanol is a kind of substance\\
as the use of a resource decreases , the length of time that resource\\ being available will increases\\
to give off means to be the source of\\
potential energy means stored energy\\
operating an automobile usually requires fossil fuels\\
amount is a property of something and includes ordered values of none / least / little / some / half / much / many / most / all\\
gasoline is a kind of chemical power source\\
to run on means to use as an energy source\\
using the fuel requires burning the fuel\\
an energy source is a kind of resource\\
using a gasoline engine requires burning gasoline\\
best means most positive\\
if the amount of electricity converted from the potential energy of an object is less than the amount of potential energy in the object , then the potential energy is converted into other energy other than electricity\\
potential energy is a measure of the energy of an object due to\\ position relative to other objects\\
oil is a kind of substance / matter\\
if an object / a substance makes something then that object / that substance is a source of that thing\\
heat means heat energy\\
an internal combustion engine is used to move the automobile\\
if something requires something else then that something requires a source of that something else\\
\textbf{Question:}\\
Which of the following is the best example of potential energy? a fallen rock OR a burning match OR a liter of gasoline OR a red-hot piece of iron.\\
\textbf{Proof:}\\
gasoline is a kind of chemical power source. We know that using a gasoline engine requires burning gasoline. Therefore,  using a gasoline engine requires burning chemical power source.\\
using a gasoline engine requires burning chemical power source. We know that gasoline is a source of chemical energy. Therefore,  using a gasoline engine requires burning chemical energy.\\
using a gasoline engine requires burning chemical energy. We know that chemical energy is a kind of potential energy. Therefore,  using a gasoline engine requires burning potential energy.\\
Answer:  a liter of gasoline\\
\textbf{Target Proof:}\\
gasoline is a source of chemical energy. We know that chemical energy is a kind of potential energy. Therefore, gasoline is a source of potential energy.
\textbf{Target Answer:} a liter of gasoline\\

\subsubsection*{Example 2 (value: -0.023)}
\textbf{Context:}\\
if heat is transferred to an object , then the object will get hot\\
heat energy is synonymous with thermal energy\\
a source of heat can be used to heat something\\
greater means higher / more in value\\
heat energy can be used to heat up an object\\
if something causes an increase in something else then that something produces that something else\\
if a body part was burned then that body part was exposed to a lot of heat energy\\
as the temperature of a chemical reaction increases , the rate of the chemical reaction will increase\\
adding heat means increasing temperature\\
fanning increases the oxygen near the fire\\
heat means heat energy\\
warm up means increase temperature\\
if something emits something else then that something increases the amount of that something else\\
amount is a property of something and includes ordered values of none / least / little / some / half / much / many / most / all\\
fire transfers heat through waves\\
if a thermal conductor / an object is exposed to a source of heat then that conductor / that object may become hot / warm\\
heat means the transfer of thermal energy\\
heating means adding heat\\
if heat is absorbed from a source then that heat source will cool\\
increase means more\\
as the amount of oxygen exposed to a fire increases, the fire will burn more\\
heat is a kind of energy\\
burning converts chemical energy into thermal energy\\
combustion means fire converts chemical energy into light energy and heat energy\\
thermal energy is a kind of energy\\
\textbf{Question:}\\
Fanning can make a wood fire burn hotter because the fanning makes the food hot enough to burn OR adds more oxygen needed for burning OR increases the amount of wood there is to burn OR provides the energy needed to keep the fire going.\\
\textbf{Proof:}\\
fanning increases the oxygen near the fire. We know that as the amount of oxygen exposed to a fire increases, the fire will burn more. Therefore,  as the amount of oxygen exposed to a fire increases, the fire will burn more quickly.\\
as the amount of oxygen exposed to a fire increases, the fire will burn more. We know that fanning increases the oxygen near the fire. Therefore,  fanning the fire increases the amount of burning woodchips.\\
as the amount of oxygen exposed to a fire increases, the fire will burn more quickly. We know that fanning the fire increases the amount of burning woodchips. Therefore,  fanning the fire increases the amount of oxygen exposed to the fire.\\
as the amount of oxygen exposed to a fire increases, the fire will burn more quickly. We know that fanning the fire increases the amount of burning woodchips. Therefore,  fanning the fire increases the amount of oxygen exposed to the fire.\\
fanning the fire increases the amount of oxygen exposed to the fire. We know that as the amount of oxygen exposed to a fire increases, the fire will burn more quickly. Therefore,  fanning the fire will cause the fire to burn more quickly.\\
\textbf{Answer:}  provides the energy needed to keep the fire going\\
\textbf{Target Proof:}\\
fanning increases the oxygen near the fire. We know that as the amount of oxygen exposed to a fire increases, the fire will burn more. Therefore, fanning can cause the fire to burn hotter.\\
\textbf{Target Answer:} adds more oxygen needed for burning\\

\subsubsection*{Example 3 (value: -0.023)}
\textbf{Context:}\\
melting point is a property of a substance / material\\
ice is cold in temperature\\
ice crystals means ice\\
definite is the opposite of variable\\
solid is a kind of state of matter\\
melting is when solids are heated above their melting point\\
an ice cube is a kind of solid\\
matter in the solid phase has definite shape\\
melting point means temperature at which a solid melts / above which a solid melts\\
state of matter means physical state\\
state means form\\
water is in the liquid state , called liquid water , for temperatures between 273 / 32 / 0 and 373 / 212 / 100 k / f / c\\
to be found in means to be contained in\\
freezing causes a solid to form\\
melting is a kind of phase change\\
state of matter at room temperature is a kind of physical property\\
a physical property is a kind of property\\
temperature / heat energy is a property of objects / weather and includes ordered values of cold / cool / warm / hot\\
liquid has a a lower melting point than solid\\
ice is colder in temperature than water\\
temperature is a measure of heat energy\\
matter in the solid phase has definite volume\\
to be made of something means to be formed of that something\\
matter in the liquid phase has definite volume\\
the properties of something can be used to identify / used to describe that something\\
\textbf{Question:}\\
Which word best describes the physical state of an ice cube? gas OR solid OR liquid OR plasma.
\textbf{Proof:}\\
an ice cube is a kind of solid. We know that solid is a kind of state of matter. Therefore,  an ice cube is in solid state.\\
\textbf{Answer:} solid\\
\textbf{Target Proof:}\\
solid is a kind of state of matter. We know that an ice cube is a kind of solid. Therefore, an ice cube is solid in its physical state.\\
\textbf{Target Answer:} solid\\

\subsubsection*{Example 4 (value: -0.038)}
\textbf{Context:}\\
oxygenated means having oxygen\\
mitochondrion provide energy to cells using oxygen and glucose\\
where something comes from is a source of that something\\
sugar is a kind of food\\
providing something means making / delivering something\\
atp is formed by the breakdown of molecules of glucose\\
if something has the ability to do something else , then that something can do that something else\\
cells are a part of the body\\
if something performs a task then that task is that something 's function\\
to provide something means to be the source of that something\\
cell division is a kind of process\\
cell activity means cell process\\
purpose means role\\
oxygen can be found in air\\
to provide means to supply\\
atp supplies chemical energy for organisms to move\\
both means two\\
to have a function is similar to to be responsible for\\
mitosis produces two cells with the same number of chromosomes as the parent cell\\
all cells take in energy\\
the function of something is what that something is used to do\\
amount is a property of something and includes ordered values of none / least / little / some / half / much / many / most / all\\
including means containing\\
energy transformation means one kind of energy changes into another kind of energy\\
a mitochondrion is a kind of cellular organelle\\
\textbf{Question:}\\
Which cellular organelle uses oxygen and glucose to provide energy to the cell? mitochondrion OR nucleus OR ribosome OR vacuole.\\
\textbf{Proof:}\\
mitochondrion provide energy to cells using oxygen and glucose. We know that a mitochondrion is a kind of cellular organelle. Therefore,  mitochondrion is the cellular organelle that provides energy to cells using oxygen and glucose.\\
\textbf{Answer:}  mitochondrion\\
\textbf{Target Proof:}\\
mitochondrion provide energy to cells using oxygen and glucose. We know that a mitochondrion is a kind of cellular organelle. Therefore, mitochondrion is the cellular organelle that uses oxygen and glucose to provide energy to cells.\\
\textbf{Target Answer:} mitochondrion\\

\subsection{EntailmentBankQA Task 2 - depth 1: SI + Halter}

A random selection of five reasoning traces on Task 2 using \textbf{SI + Halter}, these include `Unknown` examples where the model did not terminate its reasoning.

\subsubsection*{Example 1}
\textbf{Context:} \\
electric currents produce magnetic fields\\ 
knowing the properties of something means knowing information about that something\\ 
luster is a property of an object / a material and includes ordered values of dull / shiny\\ 
more likely means increased likelihood\\ 
if something is a part of something else then that something else contains that something\\ 
nitratine is always magnetic\\ 
iron is a kind of element\\ 
if something is a part of something then that something can be found in that something\\ 
cadmium is always magnetic\\ 
the metal is the core of electromagnet\\ 
iron in contact with water and oxygen will rust\\ 
the properties of something can be used to identify / used to describe that something\\ 
tungsten is a kind of metal\\ 
steel / carbon steel is always ferromagnetic / magnetic\\ 
iron is always magnetic\\ 
iron nails are made of iron\\ 
classifying means grouping objects / materials by their properties\\ 
similar means in common\\ 
a electromagnet is formed by attaching an iron nail wrapped in a copper wire to a circuit\\ 
electric currents produce magnetic fields\\ 
a magnetic field is a kind of force\\ 
nickel is always ferromagnetic / magnetic\\ 
cobalt is always ferromagnetic / magnetic\\ 
magnetic force means magnetism\\ 
the properties of something are used for describing that something\\
\textbf{Question:} \\ 
Which object most likely has magnetic properties? iron nail OR plastic clip OR rubber eraser OR wooden spoon.\\
\textbf{Proof:}\\
 iron is always magnetic. We know that iron nails are made of iron. Therefore, iron nails are magnetic.\\ 
\textbf{Answer:}\\  iron nail
\textbf{Target Proof:}\\
 iron is always magnetic. We know that iron nails are made of iron. Therefore, iron nails are magnetic.\\
\textbf{Target Answer:} iron nails are magnetic

\subsubsection*{Example 2}
\textbf{Context:} \\
celestial bodies / celestial objects are found in space\\ 
if a condition is necessary for a process , then that condition make the process possible\\ 
our sun is located at the center of our solar system\\ 
gravity causes orbits\\ 
to keep something occurring means to cause that something to occur\\ 
exerting force on an object means a force acts on an object\\ 
motion / movement means moving / to move\\ 
isaac newton discovered the theory of gravity\\ 
take place means happen\\ 
planets in the solar system orbit the sun\\ 
gravity means gravitational pull / gravitational energy / gravitational force / gravitational attraction\\ 
can be means able to be\\ 
both means two\\ 
a force is a kind of push or pull on an object\\ 
planets are found in space\\ 
requiring something means needing that something\\ 
revolving around something means orbiting that something\\ 
mars is a kind of planet\\ 
if something required by something else is in a place then that something else may be in that place\\ 
to cause means to have an effect\\ 
the sun is the strongest source of gravity in the solar system\\ 
to depend on / to rely on / to need means to require\\ 
to be used for something means to be required by that something\\ 
orbit is a kind of characteristic\\ 
due to means caused by\\
\textbf{Question:} \\ 
The force necessary to keep planets in orbit around the Sun is gravity. OR friction. OR magnetism. OR nuclear..\\
\textbf{Proof:}\\
 planets in the solar system orbit the sun. We know that gravity causes orbits. Therefore, gravity causes the planets in the solar system to orbit the sun.\\ 
\textbf{Answer:} gravity\\
\textbf{Target Proof:}\\
 planets in the solar system orbit the sun. We know that gravity causes orbits. Therefore, gravity causes planets in the solar system to orbit the sun.\\
\textbf{Target Answer:} gravity causes planets in the solar system to orbit the sun

\subsubsection*{Example 3}
\textbf{Context:} \\
  human eyes can only see light in the range of visible light\\ 
classifying is when one sorts something by kind / type\\ 
visible light is a part of the electromagnetic spectrum\\ 
both means two\\ 
if an object reflects a light of a certain color then the object appears to be that color\\ 
category means kind\\ 
made up of means contains / made of\\ 
a type is synonymous with a kind\\ 
electromagnetic energy travels as electromagnetic waves\\ 
knowing the properties of something means knowing information about that something\\ 
spectrum means a range of colors\\ 
all waves transfer / transmit energy from one place to another\\ 
the properties of something are used for describing that something\\ 
radiation means radiant energy\\ 
light means light energy\\ 
electromagnetic radiation means electromagnetic energy\\ 
light rays means light\\ 
light is a kind of nonliving thing\\ 
the properties of something can be used to identify / used to describe that something\\ 
uv light is a kind of light\\ 
the color of an object can be discovered by looking at that object\\ 
amount is a property of something and includes ordered values of none / least / little / some / half / much / many / most / all\\ 
visible light is made of lights of different colors including violet, blue, green, yellow, orange, red light\\ 
if something is a part of something else then that something else contains that something\\ 
color is a property of an object / a material\\
\textbf{Question:} \\ 
The electromagnetic spectrum is the complete range of electromagnetic radiation ranging from radio waves to gamma rays. Which type of radiation in this spectrum is associated with observable colors such as red, green, and blue? gamma rays OR X-rays OR visible light OR radio waves.\\
\textbf{Proof:}\\
 visible light is made of lights of different colors including violet, blue, green, yellow, orange, red light. We know that visible light is a part of the electromagnetic spectrum. Therefore, visible light is a part of the electromagnetic spectrum that includes violet blue green yellow orange and red light.\\ 
\textbf{Answer:}  visible light\\
\textbf{Target Proof:}\\
 visible light is made of lights of different colors including violet, blue, green, yellow, orange, red light. We know that visible light is a part of the electromagnetic spectrum. Therefore, visible light is the part of electromagnetic spectrum that has red, green, and blue light.\\
\textbf{Target Answer:} visible light is the part of electromagnetic spectrum that has red, green, and blue light

\subsubsection*{Example 4}
\textbf{Context:} \\
  to have a function is similar to to be responsible for\\ 
to enable is to help\\ 
providing support is a kind of function\\ 
a function is a kind of characteristic\\ 
a trunk is a part of a tree for supporting the tree\\ 
if an organism can do something , then that organism is able to do that something\\ 
bark is a protective covering around the trunk of / branches of a tree\\ 
purpose means role\\ 
role means function\\ 
if something has the ability to do something else , then that something can do that something else\\ 
to be used for something means to be required by that something\\ 
to be used for something means to help to do that something\\ 
a function is an activity\\ 
the function of something is what that something is used to do\\ 
a component of something means a part of that something\\ 
essential means required\\ 
ability is a property of things and includes ordered values of able / unable / can / cannot\\ 
to anchor means to provide support\\ 
role is similar to importance\\ 
a cell wall provides support / protection / shape / structure for a plant cell\\ 
something is used for that something 's purpose\\ 
if something requires something else then that something else is important to that something\\ 
helping means something can be done / completed more easily / better\\ 
being used for is similar to providing\\ 
function is similar to purpose\\
\textbf{Question:} \\ 
The main function of a tree's trunk is to provide air OR fruit OR sunlight OR support.\\
\textbf{Proof:}\\
 a cell wall provides support / protection / shape / structure for a plant cell. We know that a trunk is a part of a tree for supporting the tree. Therefore, the cell wall is the part of a plant cell that supports the plant.\\ 
the cell wall is the part of a plant cell that supports the plant. We know that bark is a protective covering around the trunk of / branches of a tree. Therefore, the cell wall is the part of a plant cell that protects the plant.\\ 
the cell wall is the part of a plant cell that supports the plant. We know that bark is a protective covering around the trunk of / branches of a tree. Therefore, the cell wall is the part of a plant cell that protects the plant.\\ 
the cell wall is the part of a plant cell that supports the plant. We know that bark is a protective covering around the trunk of / branches of a tree. Therefore, the cell wall is the part of a plant cell that protects the plant.\\ 
the cell wall is the part of a plant cell that supports the plant. We know that bark is a protective covering around the trunk of / branches of a tree. Therefore, the cell wall is the part of a plant cell that protects the plant.\\ 
the cell wall is the part of a plant cell that supports the plant. We know that bark is a protective covering around the trunk of / branches of a tree. Therefore, the cell wall is the part of a plant cell that protects the plant.\\ 
\textbf{Answer:} Unknown\\
\textbf{Target Proof:}\\
 providing support is a kind of function. We know that a trunk is a part of a tree for supporting the tree. Therefore, a function of a tree's trunk is to provide support.\\
\textbf{Target Answer:}\\ a function of a tree's trunk is to provide support

\subsubsection*{Example 5}
\textbf{Context:} \\
  metal is an electrical / electrical energy conductor\\ 
can means having the ability to\\ 
different materials usually have different properties\\ 
material means substance\\ 
to demonstrate means to exhibit\\ 
characteristic means property\\ 
iron is a kind of metal\\ 
both means two\\ 
ability is a property of things and includes ordered values of able / unable / can / cannot\\ 
motion / movement means moving / to move\\ 
if an object is made of something then the parts of that object are made of that something\\ 
similar means in common\\ 
a demonstration is similar to an example\\ 
if something has the ability to do something else , then that something can do that something else\\ 
what something is made of is a property of the organization or that something\\ 
metal is flexible\\ 
best means most positive\\ 
the properties of something can be used to identify / used to describe that something\\ 
property is similar to measurement\\ 
a wire is usually made of metal\\ 
a wire is a part of an electrical circuit for conducting electricity\\ 
a property is a kind of information\\ 
metal is a kind of substance\\ 
the properties of something are used for describing that something\\ 
an insulator is the opposite of a conductor\\
\textbf{Question:} \\ 
Flexibility is a physical property of some matter. Which of these materials best demonstrates the property of flexibility? a mirror OR a pencil OR a metal wire OR a telephone pole.\\
\textbf{Proof:}\\
 metal is flexible. We know that a wire is usually made of metal. Therefore, metal is flexible and a wire is made of metal.\\ 
\textbf{Answer:}  a metal wire\\
\textbf{Target Proof:}\\
 metal is flexible. We know that a wire is usually made of metal.\\ Therefore, a metal wire is flexible.\\
\textbf{Target Answer:} a metal wire is flexible

\section{Baseline Model Outputs}

\subsection{EntailmentWriter on ProofWriter}

Below are example \textbf{EntailmentWriter + Answer} proofs on the ProofWriter dataset. We have replaced the sentence labels with the corresponding statements from the context and have not shown the context for brevity. We see that in each case the final conclusion answers the question, but using the incorrect reasoning. We see that the model often makes up inferences that do not follow, but do help the model solve the problem. Unlike SI, whose inference step does not have access to the question, EntailmentWriter has direct access to the question and therefore has the incentive to make up inferences that help to answer the questions even if they do not follow logically.

\subsubsection*{Example 1}

\textbf{Question:} Does it imply that the statement "The dog is big" is True?\\
\textbf{Reason:}
If something eats the bear then the bear needs the lion. We know that the dog eats the bear. Therefore, the bear needs the lion.\\
If something needs the lion then the lion eats the bear. We know that the bear needs the lion. Therefore, the lion eats the bear.\\
If the lion eats the bear then the bear does not eat the dog. We know that the lion eats the bear. Therefore, the bear does not eat the dog.\\
If something is red and it does not eat the dog then the dog sees the rabbit. We know that the rabbit needs the dog and the bear does not eat the dog. {\color{red} Therefore, \textbf{the dog sees the rabbit}.}\\
If something sees the rabbit then it is not big. We know that {\color{red}\textbf{the dog sees the rabbit}}. Therefore, the dog is not big.\\

\subsubsection*{Example 2}

\textbf{Question:} Does it imply that the statement "The dog does not chase the dog" is True?\\
\textbf{Reason:}
If something chases the mouse then it is big. We know that the cow chases the mouse. {\color{red} Therefore, \textbf{the cow chases the cow}.} \\
If something chases the cow and it is big then it does not chase the dog. We know that the cow chases the mouse and {\color{red}\textbf{the cow chases the cow}. Therefore, the cow needs the dog.}\\
If something chases the cow then it needs the dog. We know that the cow chases the mouse. {\color{red}Therefore,\textbf{ the dog is nice}.}\\
If something needs the dog and it is nice then the dog is nice. We know that {\color{red}\textbf{the dog is nice}} and the cow is nice. {\color{red}Therefore, \textbf{the dog chases the cow}.}\\
If something is nice then it chases the cow. We know that {\color{red}\textbf{the dog chases the cow}. Therefore, the dog does not chase the dog.}\\

\subsubsection*{Example 3}\label{sec:EW_on_PW_cheating}

\textbf{Question:} Does it imply that the statement "The cat sees the cat" is True?\\
\textbf{Reason:}
All nice things are young. We know that The cow is nice. {\color{red}Therefore, \textbf{the cat is nice}}.
If something is young and nice then it sees the cat. We know that the {\color{red}\textbf{cat is nice}} and the cow is nice. {\color{red}Therefore, \textbf{the cat is young}}.
If something sees the cat and the cat needs the cow then {\color{red}\textbf{cat is nice}}. We know that {\color{red}\textbf{the cat is young}} and the cat needs the cow. {\color{red} Therefore, the cat sees the cat}.

\end{document}